\documentclass[a4paper,UKenglish,cleveref, autoref, thm-restate]{lipics-v2021}
\nolinenumbers
\hideLIPIcs
\usepackage[labelformat=simple]{subcaption}
\DeclareCaptionLabelFormat{subcaptionlabel}{\normalfont(\textbf{#2}\normalfont)}

\usepackage{epsfig}
\usepackage{graphicx}
\usepackage{color}
\usepackage{url}

\usepackage{amsmath}
\usepackage{amssymb}
\usepackage{xspace}
\usepackage{booktabs}
\usepackage{algorithm}
\usepackage[noend]{algpseudocode}

\newcommand {\mm}[1] {\ifmmode{#1}\else{\mbox{\(#1\)}}\fi}

\newcommand{\ignore}[1]{}

\newsavebox{\smallProofsym}                            
\savebox{\smallProofsym}                               %
{
\begin{picture}(6,6)                                   %
\put(0,0){\framebox(6,6){}}                            %
\put(0,2){\framebox(4,4){}}                            %
\end{picture}                                          %
}                                                      %

\newcommand{\Rspace}        {\mm{{\mathbb R}}}

\newcommand{\Skip}[1]       {}

\newcommand{\new}[1]{\textcolor{black}{\textbf{{#1}}}}
\newcommand{\myparagraph}[1]{\smallskip\noindent\textbf{#1}}

\title{Bregman-Hausdorff divergence: strengthening the connections between computational geometry and machine learning}

\titlerunning{Bregman-Hausdorff divergence}

\author{Tuyen Pham}{University of Florida, Gainesville, US}{tuyen.pham@ufl.edu}{}{}
\author{Hana Dal Poz Kou\v{r}imsk\'{a}}{University of Potsdam, Potsdam, Germany}{hana.dal.poz.kourimska@uni-potsdam.de}{}{}
\author{Hubert Wagner}{University of Florida, Gainesville, US}{hwagner@ufl.edu}{}{}

\authorrunning{T. Pham, H. Kou\v{r}imsk\'{a}, and H. Wagner} 


\begin{document}
\maketitle
\abstract{
The purpose of this paper is twofold. On a technical side, we propose an extension of the Hausdorff distance from metric spaces to spaces equipped with asymmetric distance measures. Specifically, we focus on the family of Bregman divergences, which includes the popular Kullback--Leibler divergence (also known as relative entropy).  \hfill \hfill\linebreak
As a proof of concept, we use the resulting Bregman--Hausdorff divergence to compare two collections of probabilistic predictions produced by different machine learning models trained using the relative entropy loss. The algorithms we propose are surprisingly efficient even for large inputs with hundreds of dimensions. \hfill \hfill\linebreak
In addition to the introduction of this technical concept, we provide a survey. It outlines the basics of Bregman geometry, as well as computational geometry algorithms. We focus on algorithms that are compatible with this geometry and are relevant for machine learning.\hfill
}
\section{Introduction}


Various kinds of data --- from sounds to images to text corpora --- are routinely represented as finite \emph{sets of vectors}. These vectors can be processed using a wide range of algorithms, often based on linear algebra. The intermediate representations as well as final outcomes of the data are, similarly, sets of vectors.

Conveniently, the above setup allows for an intuitive geometric interpretation. Indeed, it is usual to equip the vector space $\mathbb{R}^d$ in which such representations live with the Euclidean metric. Geometric objects in the geometry induced by this metric, such as the distance itself, balls and their intersections, bisectors, etc., help explain such algorithms in intuitive terms.

In recent years, however, other notions of distances have started to play an important role. One popular distance is the Kullback--Leibler divergence, often referred to as the \emph{relative entropy}. A form of this divergence is the \emph{cross entropy} loss, commonly used for training deep learning models, in particular.

Compared to the once popular \emph{mean squared error} loss (based on the Euclidean metric) the cross entropy loss (based on the Kullback--Leibler divergence) provides significantly better performance. While the Kullback--Leibler divergence is often viewed as a distance between probability vectors, it lacks standard 
features of a metric. In particular, it is typically non-symmetric and never satisfies the triangle inequality.  As such, it is less intuitive.

It may therefore be surprising that the Kullback--Leibler divergence induces a well-behaved geometry. Moreover, there exists an infinite family of distance measures, so-called \emph{Bregman divergences}, that induce similar geometries. The aforementioned Kullback--Leibler divergence is one of its most prominent members, along with the squared Euclidean distance.

There is a significant overlap between algorithms in machine learning and computational geometry. Nevertheless, computational geometry tends to focus on the Euclidean distance (and other metric distances). In contrast, the non-metric aspects of the Kullback--Leibler divergence (and other Bregman divergences) prevent computational geometry algorithms from working --- at least at the first glance. 

It turns out that several popular algorithms can be extended to the Bregman setting --- despite the lack of symmetry and triangle inequality, which are often deemed crucial.
While this is an ongoing direction, there have been efforts to extend popular algorithms to operate within this framework.

In the first part of the paper, we offer a geometric perspective on Bregman divergences. We hope this perspective will streamline further development and analysis of algorithms at the intersection of machine learning and computational geometry; in particular, in the context of data measured using relative entropy, such as probabilistic predictions returned by a classifier trained using cross entropy.

In the second part, we develop a crucial geometric tool in the context of Bregman geometry. The idea is simple: where a Bregman divergence provides a comparison between two vectors, we propose a natural way of comparing two \emph{sets} of vectors. This idea is analogous to the Hausdorff distance between two sets and we therefore call it a \emph{Bregman--Hausdorff divergence}. Interestingly, the lack of symmetry allows for several different definitions --- we select three, guided by the geometric interpretation of the original Hausdorff distance. Additionally, we propose first algorithms for computing these new divergences. These algorithms are enabled by recent development in Bregman nearest-neighbour search, and we experimentally show they are efficient in practice.

Our contribution extends the arsenal of tools capable of handling data living in Bregman geometries. One crucial example of such data is the set of probabilistic predictions of modern classifiers trained with the cross entropy loss.

\myparagraph{Paper outline.}
In the first part of the paper, we introduce concepts from information theory and a geometric interpretation for the relative entropy (Section~\ref{sec:InfoTheory}). This interpretation connects the relative entropy to a larger family of distance measures known as Bregman divergences. After a brief introduction to this family and the geometry its members induce (Section~\ref{sec:BregmanDivergence}), we explain why the asymmetry is a beneficial property in context for machine learning, and highlight computational tools that have been extended to this setting (Section~\ref{sec:BregmanAlgos}).

The second part of this paper introduces three new measurements based on Bregman divergences. We provide definitions as well as interpretations in the context of comparing sets of probability distributions (Section~\ref{sec:BHDiv}). We then provide efficient algorithms for these measurements (Section~\ref{sec:HausdorffAlgos}).
In Section~\ref{sec:Experiments}, we experimentally show that the new measurements can be efficiently computed in practical situations. In particular, we combine the theory and tools from the two previous sections to provide quantitative ways to analyze and compare machine learning models. Section~\ref{sec:Ending} concludes the paper.

\part{Survey on Bregman geometry}    
\section{Information theory and relative entropy}\label{sec:InfoTheory}
We begin by highlighting certain concepts from information theory, with the goal of providing an interpretation of the relative entropy. We will use this interpretation to develop intuition for the geometry induced by the relative entropy in Section~\ref{fig:bregman_div}. (More details on information theory can be found in~\cite{gray2013entropy, 2012elementsOfInfoTheory}.) In particular, we emphasize the inherent asymmetry of the relative entropy, which will inform our decision to focus on asymmetric versions of the Hausdorff distance later. We also provide geometric interpretation of the relative entropy. This interpretation is shared among all Bregman divergences, and will be our focus afterwards.

\myparagraph{Shannon's entropy.}
We first set up a running example to guide us through each definition in this section. Let $E_1, E_2, E_3, E_4$ be events occurring with probabilities $\tfrac{1}{2}, \tfrac{1}{4}, \tfrac{1}{8}, \tfrac{1}{8}$ respectively. We plan to transmit information on sequences of observed events. To this end we first encode each event as a finite sequence of bits, called a \emph{codeword}. We aim to minimize the expected length of a codeword with the restriction that sequences of codewords be uniquely decodable.

Consider Table~\ref{tab:ShannonEntropyExample}, in which $p_i$ is the probability of event $E_i$. The three rightmost columns provide three different codes.
\begin{table}[h!]
    \centering
    \caption{Example codes for a probability distribution on four events.}
    \begin{tabular}{ccccc}\toprule
         & Probability & Code$_1$ & Code$_2$ & Code$_3$ \\\cmidrule(lr){2-2}\cmidrule(lr){3-3}\cmidrule(lr){4-4}\cmidrule(lr){5-5}
        $E_1$ & $1/2$ & 00 & 0  & 0 \\
        $E_2$ & $1/4$ & 10 & 1  & 10 \\
        $E_3$ & $1/8$ & 01 & 01 & 110 \\
        $E_4$ & $1/8$ & 11 & 11 & 111 \\\bottomrule
    \end{tabular}
    \label{tab:ShannonEntropyExample}
\end{table}

Given a code and a probability distribution $p$, we can compute the expected code length for the transmission of information about the events. Specifically, the expected length, $\mathbb{E}[\ell_i]$, for each  code $i$ is:
\begin{align*}
    \mathbb{E}[\ell_1] &= 2\times\frac{1}{2}+2\times\frac{1}{4}+2\times\frac{1}{8}+2\times\frac{1}{8} = 2,\\
    \mathbb{E}[\ell_2] &= 1\times\frac{1}{2}+1\times\frac{1}{4}+2\times\frac{1}{8}+2\times\frac{1}{8} = \frac{5}{4},\\
    \mathbb{E}[\ell_3] &= 1\times\frac{1}{2}+2\times\frac{1}{4}+3\times\frac{1}{8}+3\times\frac{1}{8} = \frac{7}{4}.
\end{align*}
While Code$_1$ may be the most straightforward way to encode these four events, we can find a more optimized code. Although Code$_2$ has a shorter expected code length, it is not decodable. Indeed, the sequence $0111$ can describe both $E_1E_2E_4$ and $E_3E_2E_2$. In contrast, Code$_3$ is decodable and has a shorter expected code length than Code$_1$.

In his seminal paper~\cite{shannon1948mathematical}, Shannon introduced a formula to compute the lower bound for the expected length of a code for a probability distribution. Specifically, given a probability distribution $p$, the Shannon's entropy is defined as
\begin{align*}
    H(p) &= \sum_{i}p_i\log_2 \frac{1}{p_i},
\end{align*} 
with $p_i\log_2 \frac{1}{p_i} = 0$ for $p_i = 0$. 

Returning to our example: for a probability distribution $p$ from Table~\ref{tab:ShannonEntropyExample}, $H(p) = \frac{7}{4} =  \mathbb{E}[\ell_3]$.
Thus, Code$_3$ in Table~\ref{tab:ShannonEntropyExample} is the optimal way to encode events $E_1,\dots,E_4$.

\myparagraph{Cross entropy.} 
Suppose we (erroneously) assume that the probability of 
the events is given by a probability 
distribution $q$, while in reality it
is $p$. How inefficient is the code optimized for $q$ compared to the code optimized for the true distribution $p$?  

In this situation, we would assign longer codewords to less probable events, as measured by $q$. However to compute the expected code length we must use the true probabilities of events, given by $p$. 

The cross entropy is an extension of Shannon's entropy that provides a lower bound on the length of such codes: 
\begin{align*}
H(p,q) = \sum_{i} p_i\log_2 \frac{1}{q_i}.
\end{align*}

In other words, \emph{the cross entropy gives the lower bound for the expected code length for events with probabilities represented by $p$, assuming they occur with probabilities represented by $q$.}

Given distributions $p$ and $q$, their cross entropy has a geometric interpretation. It is the approximation of $H(p)$ by the best affine approximation of $H$ centered at $q$. Indeed, we can write the cross entropy as 
\begin{align*}
H(p,q)=H(q) + \langle\nabla H(q), p-q\rangle,
\end{align*} 
where $\langle\cdot, \cdot\rangle$ is the standard dot product.

\myparagraph{Relative entropy.} Relative entropy is the difference between cross entropy and entropy, $H(p,q) - H(p)$. It therefore measures the expected \emph{loss of coding efficiency} incurred by using the `approximate' probability $q$ instead of the `true' probability $p$. 

The relative entropy is often viewed as a distance measure between two probability distributions. However, unlike proper metric distances, it is \emph{generally not symmetric and does not satisfy the triangle inequality}. 

In this article, we write 
\begin{align*}
D_{KL}(p\|q) &= H(p,q)-H(p)\\
             &= \sum_{i=1}^d p_i\log_2\frac{1}{q_i} - \sum_{i=1}^d p_i\log_2\frac{1}{p_i}\\
             &= \sum_{i=1}^{d}p_i\log_2\frac{p_i}{q_i}
\end{align*}
to denote the relative entropy. We provide a further explanation of this notation at the end of this section, and expand on it in Section~\ref{sec:BregmanDivergence}.

\begin{figure}
    \centering
    \includegraphics[width=0.5\linewidth]{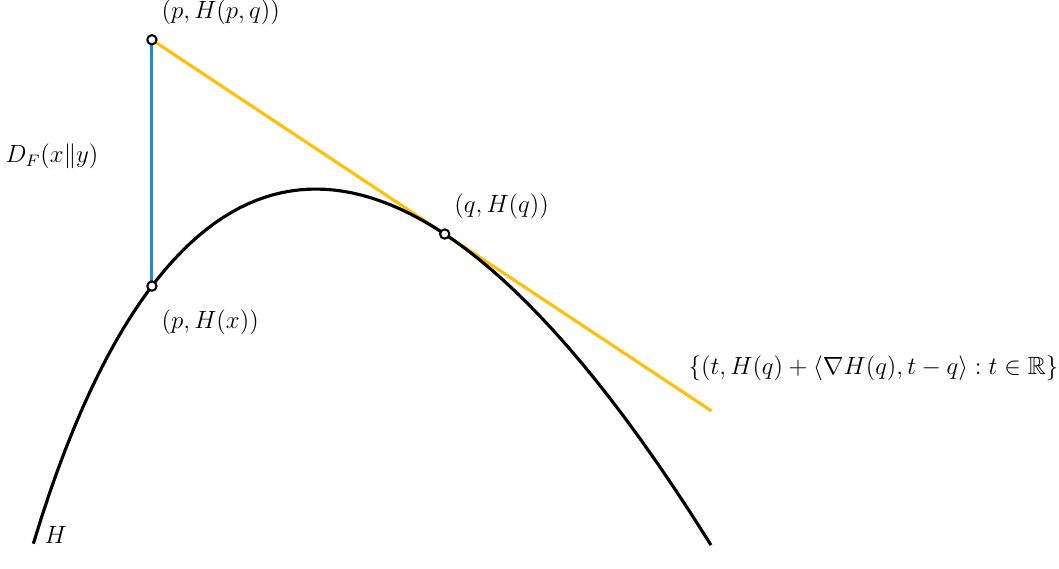}
    \caption{A visualization of the relative entropy.}
    \label{fig:relEnt}
\end{figure}

\myparagraph{Usage.} Relative entropy is often used as a loss function in machine learning models. Let us consider a multiclass classification task, with $X\subset \mathbb{R}^d$ being a data set and $Y$ the collection of correct labels encoded as probability vectors. For a model $M$ dependent on a parameter vector $\theta$ and $x\in X$, we write $M(x;\theta)$ for the probability vectors which are compared to the correct labels in $Y$. Then minimizing $\sum_{(x,y)\in X\times Y}D_{KL}(y\|M(x;\theta))$ can be interpreted as penalizing predictions that are poor approximations of the true distribution. Relative entropy has also been used in the training of Variational Autoencoders~\cite{KingmaWelling_VAEs}.

\myparagraph{Asymmetry.} The asymmetry of relative entropy has a tangible interpretation. Let $p = (\tfrac{1}{2}, \tfrac{1}{4}, \tfrac{1}{8}, \tfrac{1}{8})$ as in Table~\ref{tab:ShannonEntropyExample}, and choose $q = (\tfrac{1}{3}, \tfrac{1}{3}, \tfrac{1}{3}, 0).$ If we attempt to approximate $p$ using $q$, then $D_{KL}(p\|q) = +\infty$. This value reflects the fact that $E_4$ is an impossible event assuming $q$, and therefore no code was prepared for it (along with all other events that occur with zero probability). Hence, there exists no codeword of finite length that we could use to encode this event.

If, on the other hand, we approximate $q$ using $p$, then $D_{KL}(q\|p) \approx 0.415$. This value reflects the fact that while $E_4$ cannot occur, the coding efficiency is decreased by accounting for a codeword that is never used.  (We remark that this asymmetry also occurs when the outputs of relative entropy are finite in both directions.)

Recall that $H(p)$ is Shannon's entropy of $p$, while the cross entropy, $H(p,q)$, is an approximation of $H(p)$ based on a affine approximation at $H(q)$. Thus, the relative entropy can be interpreted as the vertical distance between the graph of this affine approximation and the graph of $H(p)$. See Figure~\ref{fig:relEnt} for illustration.

This geometric construction of relative entropy can be generalized. Indeed, the family of distance measures arising from this type of construction is known as Bregman divergences~\cite{BREGMAN1967200}. Like relative entropy, these distance measures generally lack symmetry and do not fulfill triangle inequality. The relative entropy is a member of this family and is often referred to as the Kullback--Leibler ($KL$) divergence in this context. We will introduce this family next.

\section{Background on Bregman geometry}\label{sec:BregmanDivergence} 
In this section, we introduce Bregman divergences~\cite{BREGMAN1967200}, portray the $KL$ divergence as an important instance, and give an overview of Bregman geometry. We will put emphasis on the geometry induced by the $KL$ divergence.

\myparagraph{Bregman divergences --- definition and basic properties.}
Each Bregman divergence is generated by a function of Legendre type. 
Given an open convex set $\Omega\subseteq\mathbb{R}^d$, a function $F:\Omega\to \mathbb{R}$ is \new{of Legendre type}~\cite{bauschke1997legendre, Rockafellar+1970} if $F$ is:
\begin{enumerate}
    \item differentiable;
    \item strictly convex;
    \item if the boundary $\partial\Omega$ of $\Omega$ is nonempty, then $\lim\limits_{x\to \partial\,\Omega}\|\nabla F(x)\|=\infty$.    
\end{enumerate}
The last condition is often omitted, but it enables a correct application of the \emph{Legendre transformation}. It is a useful tool coming from convex optimization~\cite{Rockafellar+1970}, which we will review later.

\begin{figure}[h!]
    \centering
    \includegraphics[width = .65\linewidth]{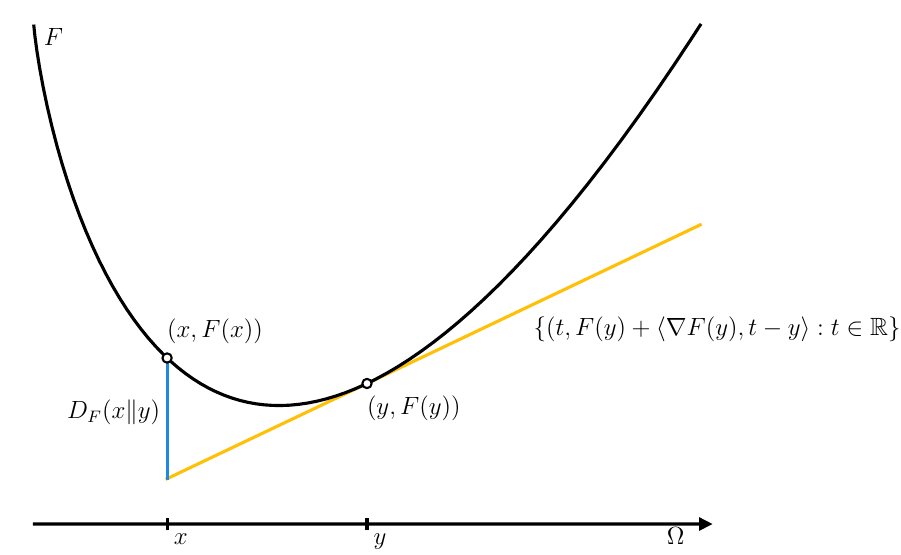}
    \caption{Visualization of a Bregman divergence formula for a one-dimensional domain.
    }
    \label{fig:bregman_div}
\end{figure}

Given a function $F$ of Legendre type, the \new{Bregman divergence}~\cite{BREGMAN1967200} generated by $F$ is the function 
\begin{align*}
    D_{F}:\Omega\times\Omega\to [0,\infty], \qquad D_F(x\|y) = F(x) - (F(y) + \langle\nabla F(y), x-y\rangle).
\end{align*}
In other words, the divergence in the \new{direction} from $x$ to $y$ is the difference between $F(x)$ and the best affine approximation of $F$ at $y$, evaluated at $x$. 
We illustrate this in Figure~\ref{fig:bregman_div}.
The construction also mirrors the geometric interpretation of the $KL$ divergence from Figure~\ref{fig:relEnt}.

Just like metrics, Bregman divergences are always non-negative: in fact, $D_F(x\|y)\geq 0$, with equality if and only if $x=y$. 
Unlike metrics, however, they are generally not symmetric, and do not satisfy the triangle inequality.
To emphasize the lack of symmetry, it is customary to write $D_{F}(x\|y)$ for the divergence computed from $x$ to $y$.

\myparagraph{Relative entropy as a Bregman divergence.}
Given the function $ F : \Rspace_+^d \rightarrow \Rspace$,  $F(x) = -E(x) = \sum_{i} x_i \log_2 x_i $, we have $ \nabla F(x) = \frac{1}{\log(2)}(1 + \log x_1, \dots, 1 + \log x_d) $.  Substituting in the formula for the Bregman divergence,
we obtain
\begin{align*}
D_{KL}(x \| y) & = D_{-E}(x \| y)   \\
 &= \sum_{i=1}^{d} x_i \log_2 x_i - \sum_{i=1}^{d} y_i \log_2 y_i - \sum_{i=1}^{d} \frac{1}{\log(2)}(1 + \log y_i)(x_i - y_i)\\
&= \sum_{i=1}^{d} x_i \log_2 x_i - x_i \log_2 y_i + \frac{y_i - x_i}{\log(2)}.
\end{align*}

This divergence is often called a \new{generalized Kullback--Leibler divergence}. Restricting it to the probability simplex, 
\begin{align*}
    \triangle^{d-1} = \{x\in \mathbb{R}^{d}_{+}\,:\,\sum_{i=1}^{d}x_i = 1 \},
\end{align*}
we obtain a divergence which coincides with the relative entropy defined in the previous section. Following statistics and information theory literature, we will refer to it as the Kullback--Leibler divergence and denote it by $D_{KL}$. One can easily check that the generator of this divergence is indeed a function of Legendre type, also when restricted to the probability simplex.

\myparagraph{Other Bregman divergences.}
Among other examples of Bregman divergences, the most prominent one is the \new{squared Euclidean distance (SE)}, generated by the square of the Euclidean norm:
\begin{align*}
    \Rspace^d\to \Rspace, \qquad x\mapsto \sum_{i=1}^dx_i^2,\\
    D_{SE}(x\|y) = \sum_{i=1}^d (x_i-y_i)^2.
\end{align*}
We remark that restricting the domain to any bounded subset of $\mathbb{R}^d$ violates Condition 3 for a Legendre type function. Consequently, under such a restriction the square of the Euclidean distance does not fulfill the above definition of a Bregman divergence. 
\begin{figure}[t]
    \centering
    \includegraphics[scale=0.45]{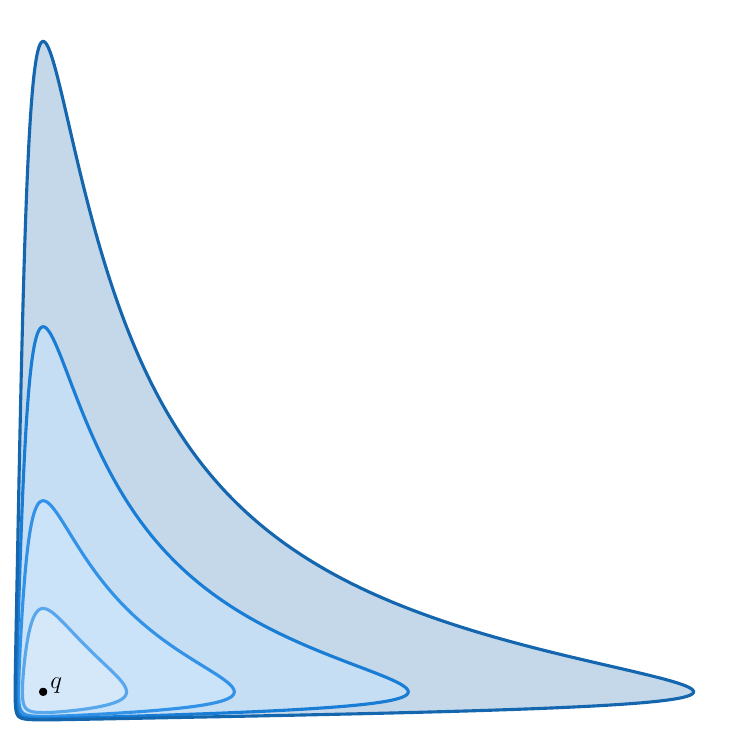}
    \includegraphics[scale=0.45]{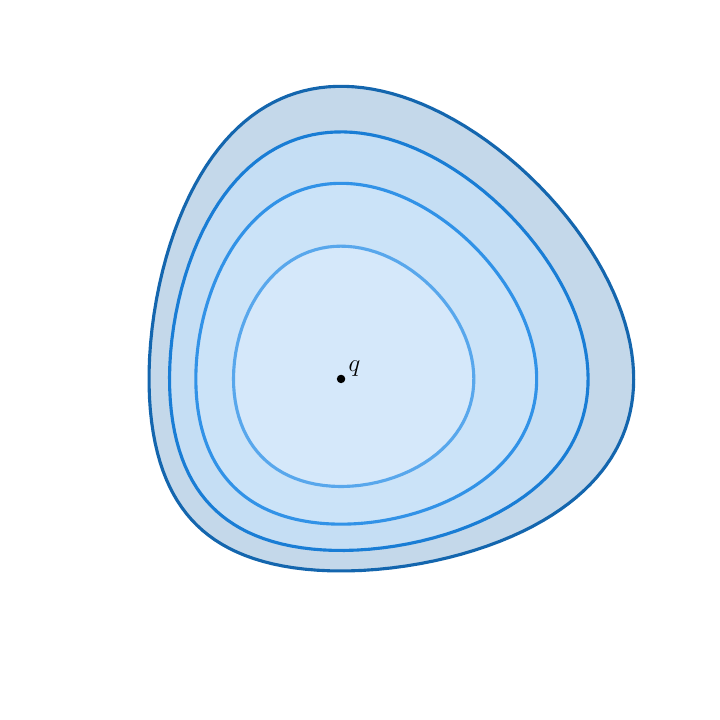}
    \caption{Left: concentric primal Itakura--Saito balls. Right: concentric primal generalized Kullback--Leibler balls.}
    \label{fig:GKL_IS_Open_Ball}
\end{figure} 

The \new{Itakura--Saito (IS) divergence}~\cite{do2002wavelet}, generated by the function
\begin{align*}
    \Rspace^d_+\to \Rspace, \qquad x\mapsto-\sum_{i=1}^d \log x_i,\\
    D_{IS}(x\|y) = \sum_{i=1}^d \frac{x_i}{y_i}-\log\frac{x_i}{y_i}-1,
\end{align*} 
has seen success as a loss function in machine learning models analyzing speech and sound data~\cite{itakura1968analysis}. 

All of the above divergences are 
often classified as \new{decomposible Bregman divergences}, since each of them decomposes as a sum of 1-dimensional divergences.

The \new{Mahalanobis divergence}~\cite{mahalanobis1936generalised} is commonly used in statistics as a distance notion between probability distributions~\cite{MahalanobisStandardUses}.
It is generated by the convex quadratic form 
\begin{align*}
    \Rspace^d\to \Rspace,& \qquad x\mapsto x^{\top}Qx,\\
    \Rspace^d\times\Rspace^d\to [0,\infty],& \qquad (x,y)\mapsto (x-y)^\top Q (x-y),
\end{align*}
where $Q$ is the inverse of a variance-covariance matrix. Unlike the other examples, it is generally not a decomposable divergence.
The Mahalanobis divergence has also seen success in machine learning --- examples include supervised clustering~\cite{MahalanobisClustering} and the classification of hyperspectral images~\cite{MahalanobisClassification}. 

\myparagraph{Bregman Geometry.} Similarly to a metric, each Bregman divergence induces a geometry. We provide a brief overview of some key objects and features of Bregman geometries, and provide their information theoretic interpretations in the case of the $KL$ divergence. 

\begin{figure}[t!]
    \centering
    \includegraphics[width = .49\textwidth]{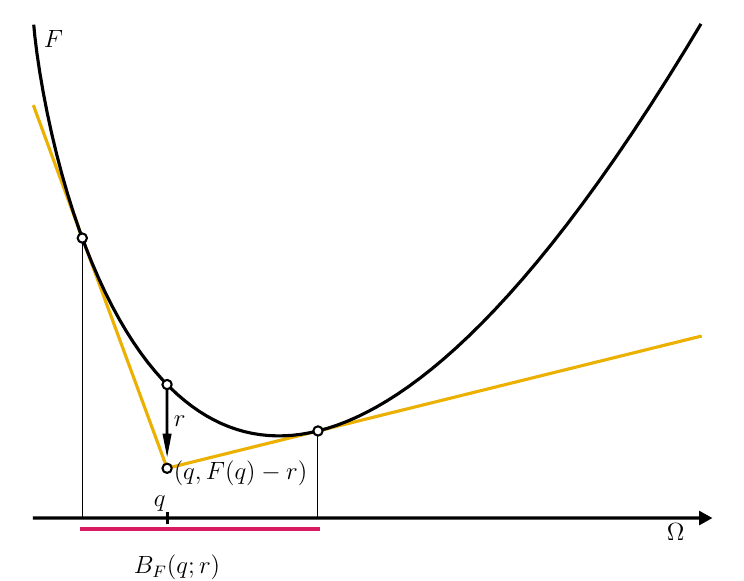}
    \includegraphics[width = .49\textwidth]{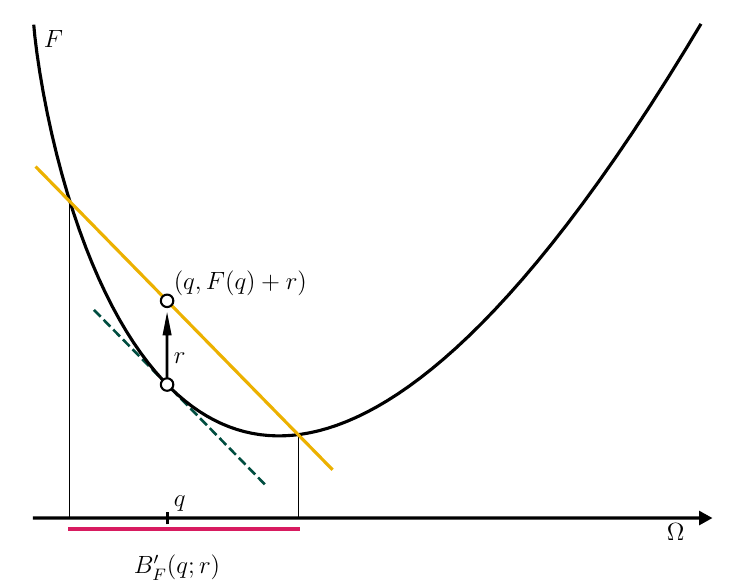}
    \caption{Geometric interpretation of primal (left) and dual (right) Bregman balls in dimension one.}
    \label{fig:PrimalDualBall_def}
\end{figure}

A fundamental object in geometry is the ball. Due to the asymmetry of Bregman divergences, one can define two types of Bregman balls~\cite{BregmanBallDef}. The \new{primal Bregman ball} of radius $r\ge 0$ centered at the point $q\in\Omega$, is defined as
\begin{align*}
    B_F(q;r) = \{y\in\Omega\,:\,D_F(q\|y) \leq r\},
\end{align*}
and is the collection of those points whose divergence \emph{from} $q$ does not exceed $r$. 
See~\cref{fig:GKL_IS_Open_Ball} for various illustrations. 
Primal Bregman balls have a particularly nice geometric interpretation: given a light source at point $(q, F(q)-r)$, the primal ball $B_F(q;r)$ is the \emph{illuminated} part of the graph of $F$, projected vertically onto $\Omega$. 
We illustrate this in Figure~\ref{fig:PrimalDualBall_def}, on the left.

The \new{dual Bregman ball} of radius $r\geq0$ centered at the point $q\in\Omega$, is defined as
\begin{align*}
    B_F'(q;r) = \{y\in\Omega: D_F(y\|q)\leq r\},
\end{align*}
and is the collection of points whose divergence \emph{to} $q$ does not exceed $r$.
The dual ball has a geometric interpretation similar to the primal ball. 
We first shift the tangent plane of the graph of $F$ at $(q, F(q))$ up by $r$. 
The dual Bregman ball $B'_F(q;r)$ is the portion of the graph of $F$ below this plane, projected vertically onto $\Omega$.
We illustrate this in Figure~\ref{fig:PrimalDualBall_def}, on the right.

As seen in Figure~\ref{fig:GKL_IS_Open_Ball}, and observed in~\cite{Bregman_Voronoi}, primal Bregman balls can be non-convex when viewed as a subset of Euclidean space. In contrast, dual balls are always convex since $D_F(x\|y)$ is convex in $x$.

It may be tempting to consider the two geometries induced by a Bregman divergence to and from a point, separately. 
However, there is a strong connection between the two, given by the Legendre transformation~\cite{Rockafellar+1970}.

\myparagraph{Legendre transform.} The \new{Legendre transform} $F^*$ of a function $F$ of Legendre type is defined on a conjugate domain 
\begin{align*}
    \Omega^* = \left\{ \nabla F(x)\mid x\in\Omega \right\}
\end{align*}
as 
\begin{align*}
    F^*(x^*) = \sup_{x\in \Omega}(\langle x,x^*\rangle - F(x)).
\end{align*}

This construction induces a canonical map 
\begin{align*}
    \Omega\to \Omega^*, \qquad x\mapsto x^* = \nabla F(x).
\end{align*}
It turns out that $F^*$ is a convex function of Legendre type, and thus it also generates a Bregman divergence, $D_{F^{*}}$. 
This divergence satisfies
\begin{align*}
    D_{F^{*}}(p*,q*) = D_F(q, p),
\end{align*}
implying that the function mentioned above maps primal balls in $\Omega$ to dual balls in $\Omega^*$~\cite{Bregman_Voronoi}.

\myparagraph{Chernoff point.} Another connection between the two geometries is given by the \new{Chernoff point}~\cite{ChernoffInfo}. Following~\cite{EdViWa17}, we say a set $X\subset\Omega$ has an \new{enclosing sphere}, if there exists a dual Bregman ball containing $X$. The enclosing sphere is then the boundary of this ball.

Moreover, every finite set $X\subset \Omega$ has a unique smallest enclosing Bregman sphere~\cite{ChernoffCompute, BregmanBallDef, EdViWa17}.
The center of this smallest enclosing sphere is known as the Chernoff point for $X$.

We are interested in a simple situation in which the set $X$ consists of two points only, $X = \{p,q\}$. In this case the Chernoff point minimizes the divergence $D_F(p\|c)$ subject to $D_F(p\|c)=D_F(q\|c)$; one can view it as lying midway between $p$ and $q$ with respect to the chosen Bregman divergence.
Indeed, in the squared Euclidean case, it is the usual midpoint, $\frac{p+q}{2}$.

\begin{figure}
    \centering
    \includegraphics[width=0.5\linewidth]{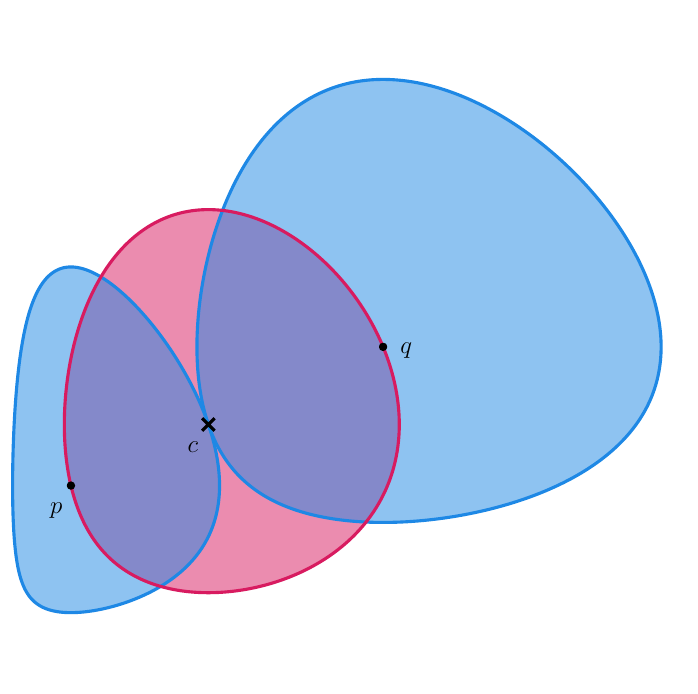}
    \caption{In blue (light): Primal $KL$ balls with centers $x$ and $y$ intersect at the Chernoff point $c$. In magenta (dark): A dual $KL$ ball of radius $D_{KL}(x\|c)$ is drawn about $c$.}
    \label{fig:chernoff_pt}
\end{figure}

The Chernoff point can be visualized: for each point $p\in X$, consider the primal ball $B_F(p;r)$ growing about $p$, parameterized by $r \ge 0$. 
Then the Chernoff point is the point where all the balls $B_F(p;r)$ intersect for the first time.
In the case when $X = \{p,q\}$, let us denote this radius by $r^*$, and the Chernoff point by $c$.
Then $c$ lies on the boundary of both $B_F(p;r^*)$ and $B_F(q;r^*)$, and $p$ and $q$ lie on the boundary of $B_F'(c;r^*)$.  
A visualization of this interaction can be seen in Figure~\ref{fig:chernoff_pt}.

We refer to works by Nielsen~\cite{ChernoffInfo, NielsenChernoffRevisit} for more information on Chernoff points and their applications.


\myparagraph{Information-theoretic interpretation of the geometric objects stemming from the $KL$ divergence.} In the language of information theory, the primal and dual $KL$ balls have the following interpretation: Let $p\in\triangle^{d-1}$ be a probability vector, and $r \ge 0$ a radius, expressed in bits. Note that, since the $KL$ divergence measures the \emph{expected} efficiency loss in bits, the radius $r$ can indeed be of any (non-negative) real value, and is not limited to being an integer. 
The primal ball $B_{KL}(p;r)$ contains all probability vectors $q$ that can \textit{approximate} $p$ with expected loss of at most $r$ bits. 
In contrast, the dual ball $B_{KL}'(p;r)$ contains all probability vectors $q$ that can be \textit{approximated by} $p$ with an expected loss of at most $r$ bits. 
Consequently, the Chernoff point for two probability vectors $p$ and $q$ is the vector that approximates \emph{both} $p$ and $q$ with the least loss of expected efficiency (as usual counted in bits).

\section{Algorithms in Bregman geometry}\label{sec:BregmanAlgos}
The development of algorithms for Bregman divergences is relatively young. We survey computational geometry algorithms that were adapted to the Bregman setting from the common Euclidean and metric settings, and highlight the necessary modifications.

\myparagraph{$k$-means clustering.} The first algorithm to be adapted was the $k$-means clustering algorithm. It was originally proposed by Lloyd~\cite{LloydKMeans} in 1957, and worked with the Euclidean distance.  It was extended to arbitrary Bregman divergences by Banerjee and collaborators~\cite{JMLR:v6:banerjee05b} in 2005. 

In Euclidean space, the $k$-means clustering partitions a dataset into $k$ clusters. 
Each cluster is identified by a unique point, called a cluster centroid. Here, the centroid is chosen as the arithmetic mean of all the data points in the cluster. This choice is motivated by the fact that the mean is the unique point that minimizes the sum of the squares of the Euclidean distances (i.e. the variance) to all data points in a given cluster.

\begin{figure}
    \centering
    \includegraphics[width=0.45\linewidth]{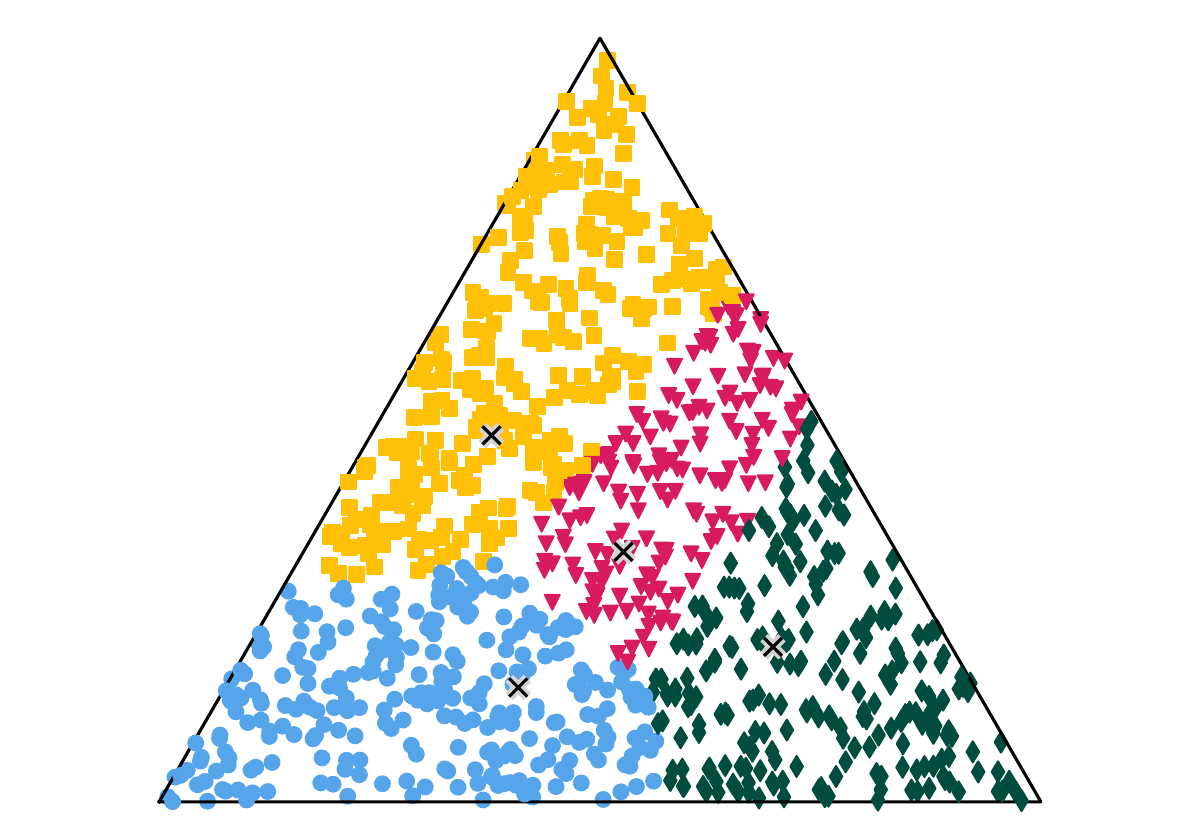}
    \includegraphics[width=0.45\linewidth]{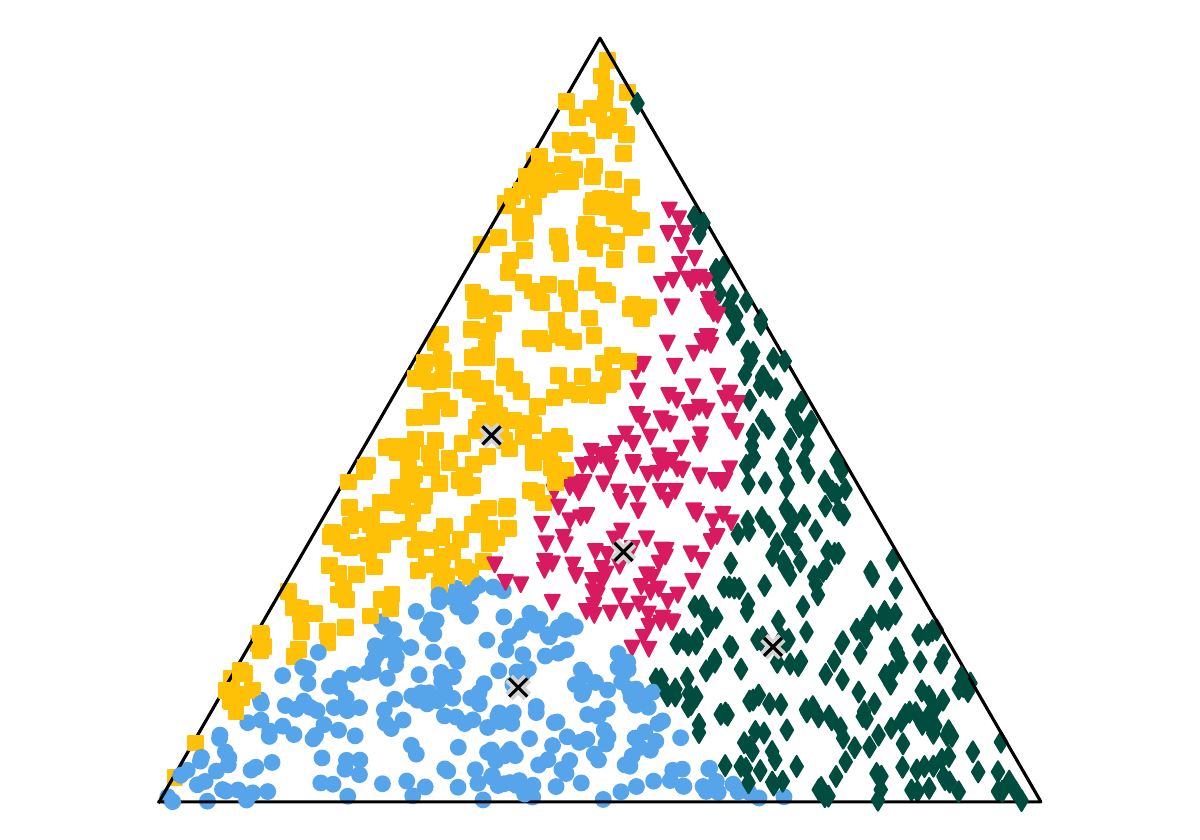}
    \caption{Clustering assignments for $k$-means with four centroids on $\triangle^2$. Left is with squared Euclidean distance, right is with $KL$ divergence. Centroids are denoted by $\times$'s.}
    \label{fig:kMeansAssignments}
\end{figure}

Surprisingly, the $k$-means algorithm still works when the square of the Euclidean distance is replaced with a Bregman divergence computed \textit{from} data points. Indeed, Banerjee and collaborators show that for $X=\{x_i\}_{i=1,2,\dots,d}\subset \Omega$, the sum $\sum_{i=1}^d D_F(x_i\|p)$ is uniquely minimized by point $p$ being the arithmetic mean of $X$ --- independently of the choice of a Bregman divergence. Consequently, apart from the requirement of computing the divergence towards the centroid, the original algorithm works in the Bregman setting without modification. This remarkable fact was the first hint that other geometric algorithms may work in the Bregman setting. (See Figure~\ref{fig:kMeansAssignments} for a comparison of $k$-means clustering in different geometries.)

\myparagraph{Voronoi diagrams.} 
Voronoi diagrams were first formally defined for two and three dimensional Euclidean space by Dirichlet~\cite{DirichletTesselation} in 1850, and for general Euclidean spaces by Voronoi~\cite{Voronoi1, Voronoi2} in 1908. Given a finite set $S\subset\mathbb{R}^d$ of points, called \new{sites}, we partition the space $\mathbb{R}^d$ into \new{Voronoi cells}, such that the Voronoi cell of $s \in S$ contains all points in the space for which $s$ is the closest site. More formally: $\{x\in \mathbb{R}^d\,:\,d(x,s)\leq d(x,y) \,\forall\, y\in S\}$.

\begin{figure}
    \centering
    \includegraphics[width = .3\linewidth]{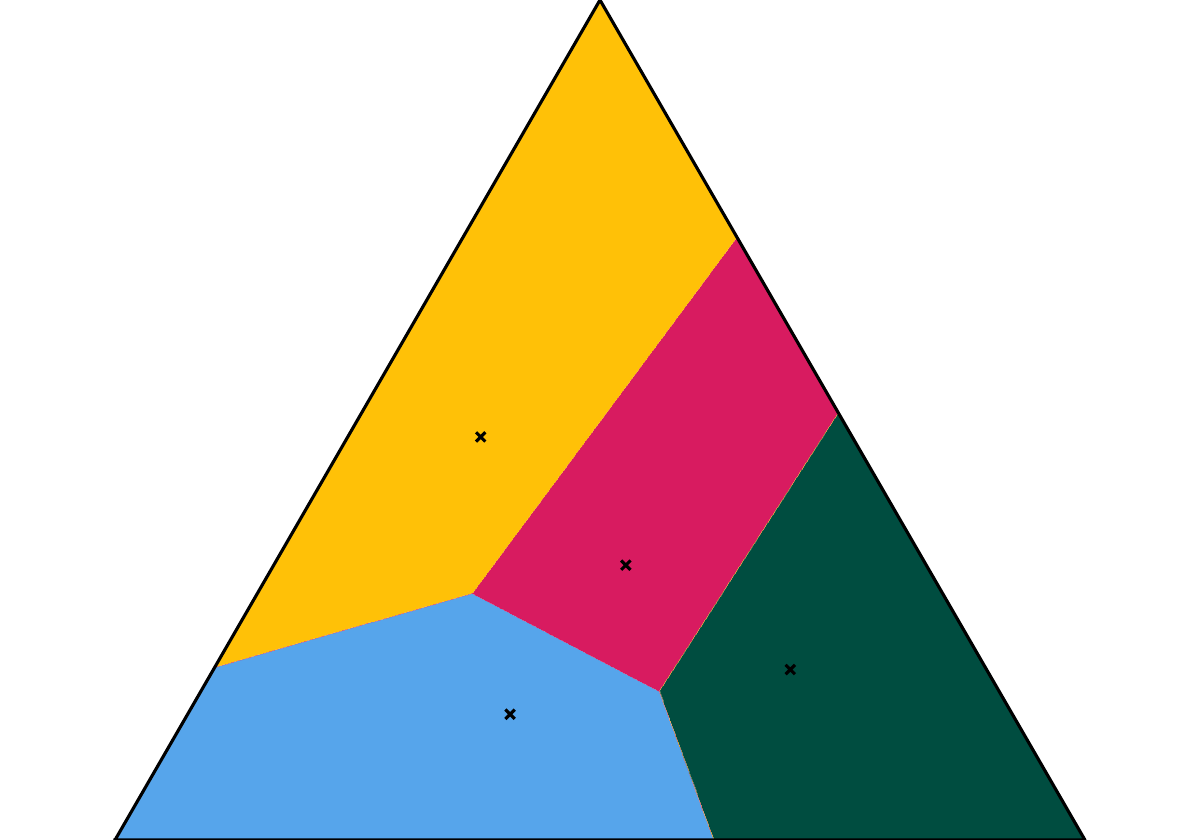}
    \includegraphics[width = .3\linewidth]{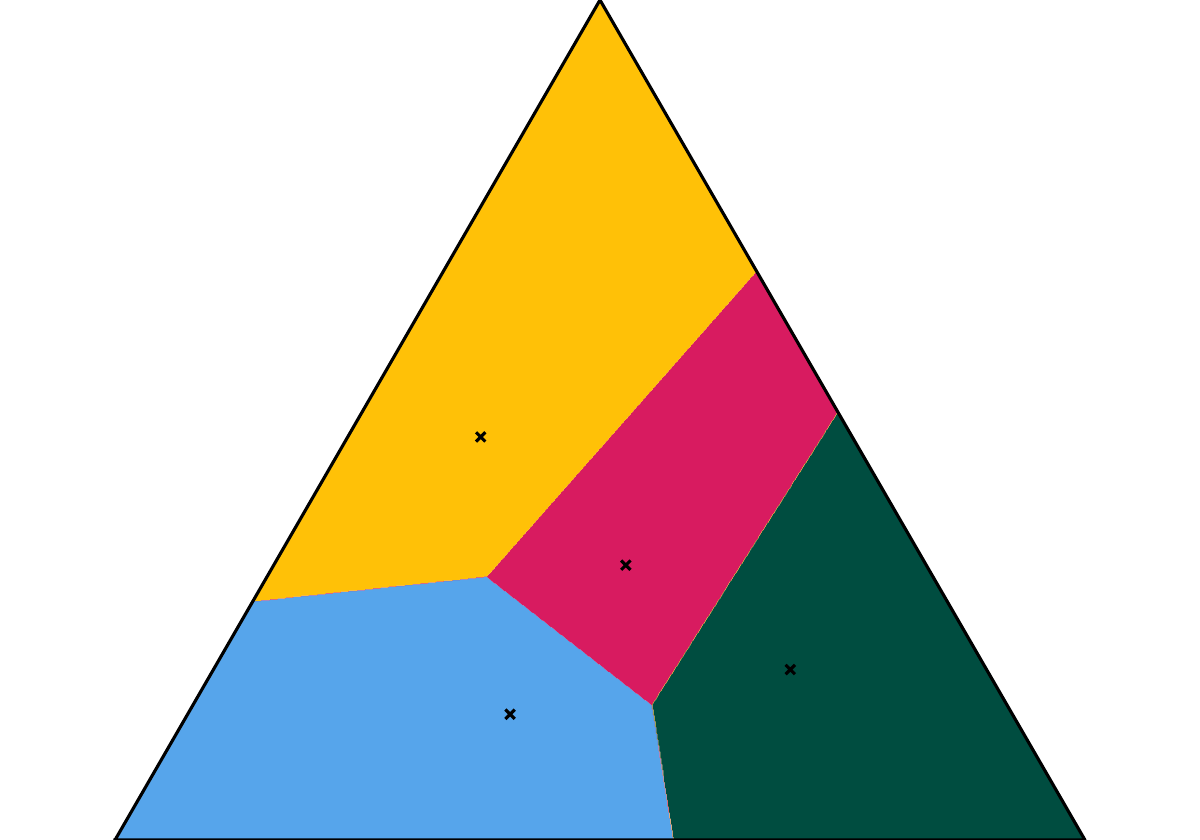}
    \includegraphics[width = .3\linewidth]{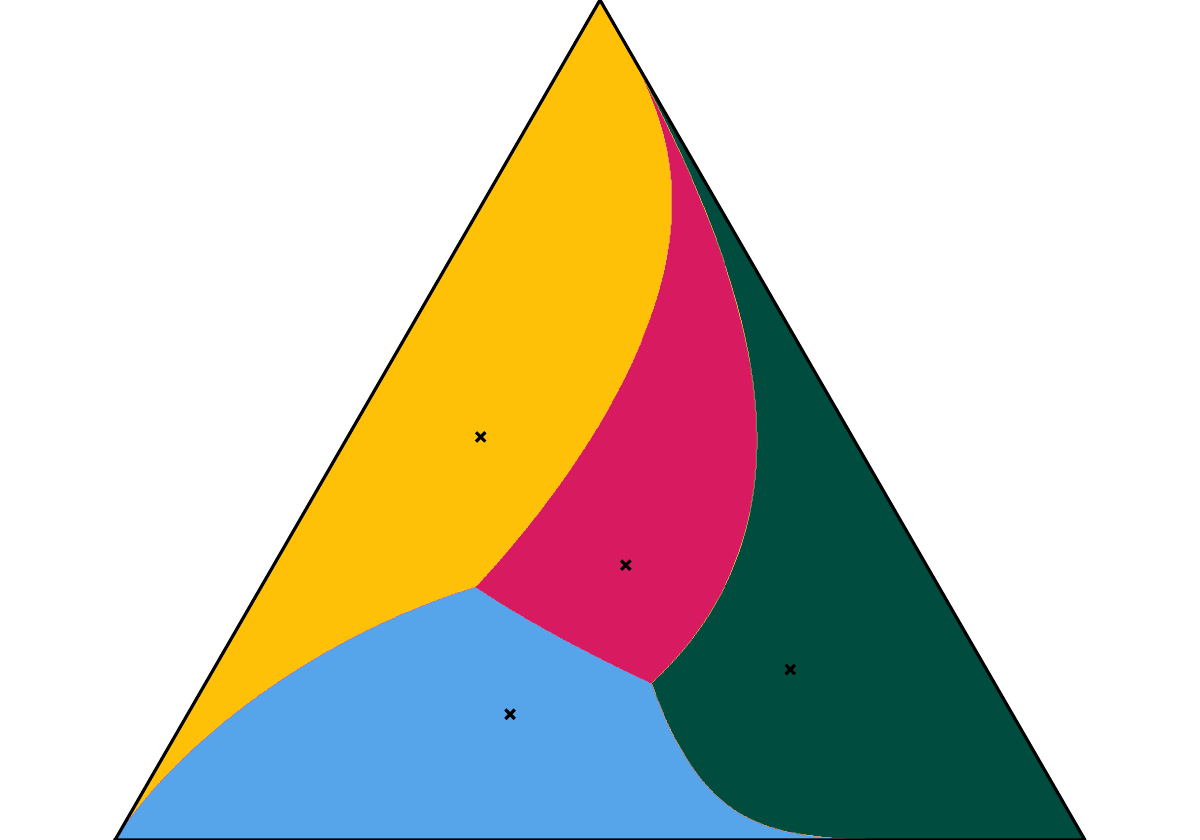}
    \caption{Voronoi diagrams on $\triangle^2$ fixed sites $S$. The left is computed with the Euclidean distance. The middle and right images compute nearest site with respect to the $KL$ divergence, the middle \textit{to} each site and the right \textit{from} each site.}
    \label{fig:L2_Linf_Voronoi}
\end{figure}

This definition for Voronoi diagrams was extended to the Bregman setting by Boissonnat, Nielsen, and Nock~\cite{Bregman_Voronoi} in 2007. In the Euclidean space, Voronoi cells are convex polytopes (possibly unbounded). However, when another distance measure is used, the shape of these cells can change drastically. See Figure~\ref{fig:L2_Linf_Voronoi} for an illustration of Voronoi diagrams for various distance measures. 

In the Euclidean case, the bisector of a pair of sites is a hyperplane, and the Voronoi cells arise as intersections of the resulting half-spaces. One method of computing these cells is by Chazelle's half-space intersection algorithm~\cite{ChazelleHalfSpaces}. In the Bregman version, Boissonnat, Nielsen, and Nock show that calculating the divergence \textit{to} the sites yields Bregman bisectors that are also hyperplanes. Chazelle's algorithm can therefore be used to compute the Bregman Voronoi diagram. However, when computing the divergence \textit{from} a site, the Bregman bisectors are more general hypersurfaces. Consequently, the Bregman Voronoi cells may have curved faces. As a side note, the Legendre transform can be used to handle this harder scenario by mapping the input to the conjugate space, performing the computations there using hyperplanes, and mapping the result back.

\myparagraph{$k$-Nearest neighbour search.} In a metric space $(M,d)$, given a finite subset $S\subset M$ and a query $q\in M$, a $k$-nearest neighbour search finds the $k$ elements of $S$ closest to $q$. A common strategy to answer these queries is to spatially decompose the set $S$.  This decomposition is often encoded as a tree data structure. This type of data structures includes ball trees~\cite{omohundro1989five}, vantage point trees~\cite{Yianilos1993}, and Kd-trees~\cite{Bentley1975MultidimensionalBS}. Collectively, these are referred to as metric trees --- although many of them can be extended to the non-metric setting of Bregman divergences.

In the Bregman setting, due to the asymmetry, we consider two types of nearest neighbour search. Namely, $\arg\min_{s\in S}D_F(q\|s)$ and $\arg\min_{s\in S}D_F(s\|q)$. Queries are performed by searching subtrees while updating a nearest neighbour candidate. Subtrees may be ignored (pruned) to accelerate the search if certain conditions, which depend on the type of tree, are met. 



In 2008, Cayton extended the concept of Ball trees from metric spaces to the Bregman setting, and created software for Bregman ball trees~\cite{CaytonBBTrees}. The software works for the squared Euclidean distance and the $KL$ divergence, with experimental support for the $IS$ divergence. Cayton's Bregman Ball trees are constructed with the help the aforementioned Bregman $k$-means clustering. The pruning decision involves a projection of a candidate point onto the surface of the Bregman ball. This is done via a bisection search.

Nielsen, Piro, and Barlaud improved on Cayton's Ball tree algorithm~\cite{Bregman_Ball_Trees}. In particular, they improved the construction by altering the initial points for the Bregman $k$-means algorithm, and introduced a branching factor to allow for more splits at each internal node. They also implemented a priority-based traversal instead of the standard recursive ball tree traversal. The same authors also adapted another data structure called a \new{vantage point tree} in 2009~\cite{Bregman_Ball_Trees, 5202635}. Specifically, they replaced the metric ball with a dual Bregman ball, and in the pruning decision, they performed a bisection search to check intersections of Bregman balls.

Kd-trees were introduced by Bentley~\cite{Bentley1975MultidimensionalBS} in 1975, and extended to Bregman divergences by Pham and Wagner~\cite{BregmanKdTrees} in 2025. Unlike ball trees and vantage point trees, the construction of the Bregman Kd-tree is independent of the Bregman divergence. Indeed, the choice of divergence can be deferred to the time of performing each query. The query algorithm, surprisingly, is the same as in the Euclidean case. For decomposable Bregman divergences in particular, it allows each pruning decision to be made in effectively $O(1)$ time, independently on the dimension of the data.

Cayton's Bregman ball trees algorithm is specialized for the $KL$ divergence, with experimental implementation for the $IS$ divergence. Nielsen, Piro, and Barlaud's implementations show results for the $KL$ and $IS$ divergences. However, for both Bregman ball trees and vantage point trees, adding implementations for new divergences is nontrivial. In contrast, although Kd-trees work for a subfamily of Bregman divergences, a further extension of the implementation to new Bregman divergences is straightforward.

Apart from the data structures and algorithms described above, other exact searches that have been extended to work in the Bregman setting. These include R-trees and VA-files (extended by Zhang and collaborators~\cite{RTreesVAFiles} in 2009), and BrePartition (introduced by Song, Gu, Zhang, and Yu~\cite{song2020brepartition} in 2020).  The listed algorithms provide exact nearest neighbour queries in the Bregman setting~\cite{nmslib}. We remark that there exist other nearest neighbour algorithms that work in the Bregman setting~\cite{nmslib, hnswg_search}. However, these algorithms focus on approximations, without guarantees to find the exact nearest neighbours.

\myparagraph{Computational topology.}
Topological concepts, such as homology groups, have been imported into computational geometry. One key concept is \new{persistent homology}~\cite{edelsbrunner2000topological}, a stable geometric-topological 
descriptor of data, including point cloud data. It is the basis of the field called \new{topological data analysis}~\cite{Car09}.

Building on geometric results of Boissonnat, Nielsen, and Nock~\cite{Bregman_Voronoi},
Edelsbrunner and Wagner~\cite{DBLP:journals/corr/EdelsbrunnerW16} extended concepts of computational topology to the Bregman setting in 2017. 
In particular, basic concepts that allow for computation of persistent homology were 
extended to the Bregman setting. These include generalizations of the alpha and \v{C}ech constructions~\cite{edelsbrunner2010computational}. One key result is the proof of contractibility of 
nonempty intersection of Bregman balls. Intuitively speaking, this result ensures that these constructions correctly capture the topology of data. More recent work focuses on implementation and experimental aspects~\cite{InformationSpaceEdelsbrunnerOlsbockWagner}.

\part{New concept: Bregman--Hausdorff divergence}
The development of computational tools for Bregman divergences motivates the extension of other geometric concepts from metric spaces to the Bregman setting. In the following sections we concentrate on one of the most commonly used --- the \emph{Hausdorff distance}. In short: we aim to compare \emph{two sets of vectors} embedded in a Bregman geometry.

In Section~\ref{sec:BHDiv}, we recall the definition of the Hausdorff distance and some of its properties. 
We then introduce the extension to the Bregman setting. Here, we offer two separate variants: the Bregman--Hausdorff and Chernoff--Bregman--Hausdorff divergence. Finally, we offer an interpretation of both divergences based on the $KL$ divergence, through the lens of information theory.
In Sections~\ref{sec:HausdorffAlgos} and~\ref{sec:Experiments}, we demonstrate how nearest neighbour algorithms can be used to compute the two divergences. 

We hope that extending the basic concept to the Bregman setting may also open door for further development. This would not be unprecedented: as mentioned above, the  development of Bregman $k$-means enabled the development of efficient Bregman Ball trees and Bregman vantage point trees.

\section{Bregman--Hausdorff divergence}\label{sec:BHDiv}
Hausdorff distance is a very natural concept: introduced by Hausdorff~\cite{hausdorff1914grundzüge} in 1914, it has since become the standard distance measure for comparing sets of points, used ubiquitously across multiple fields of mathematics.

Recently, Hausdorff distance has also been used in applications. Indeed, it is a natural choice whenever two shapes need to be compared. For example, in computer vision, Hausdorff distance has been implemented as a measurement for the largest segmentation error in image segmentation~\cite{HausdorffVision, reducinghausdorffdistancemedical}; and to compare 3D shapes, such as meshes~\cite{MetroMeshHaus, GutheFastApxHuas}.

We start this section by providing the definition of the Hausdorff distance in a metric space. We then extend this concept to the Bregman setting. The inherent asymmetry of Bregman divergences leads to a number of distinct definitions, which would all coincide in the metric setting. In devising the definitions, we are guided by the geometric and information-theoretical considerations.  We elaborate on situations in which each of these new definitions finds a natural application.


\myparagraph{Hausdorff distance in metric spaces.}\label{sec:HausdorffReview}
Given two sets $P$ and $Q$ in a metric space $(M,d)$, the \new{one-sided Hausdorff distance} from $P$ to $Q$ is defined as 
\begin{align*}
    d(P,Q) = \sup_{p\in P}\inf_{q\in Q}d(p,q).
\end{align*}
Similarly to the Bregman divergence, this measurement is not symmetric: $d(P,Q)\neq d(Q,P)$. 
The \new{Hausdorff distance} between $P$ and $Q$ is the symmetrization of the two one-sided Hausdorff distances, and is given by the maximum, \begin{align*}
    H_d(P,Q) = \max\{d(P,Q), d(Q,P)\}.
\end{align*}
Equivalently, the one-sided Hausdorff distance can be defined using a so-called \new{thickening}. A thickening --- sometimes also called an offset --- of the set $Q$ of size $r$ consists of all those points in the ambient space $M$, whose distance to $Q$ is at most $r$. In other words, it is the union of all balls of radius $r$ centered at a point in $Q$. 

\begin{figure}
    \centering
    \includegraphics[width=0.45\linewidth]{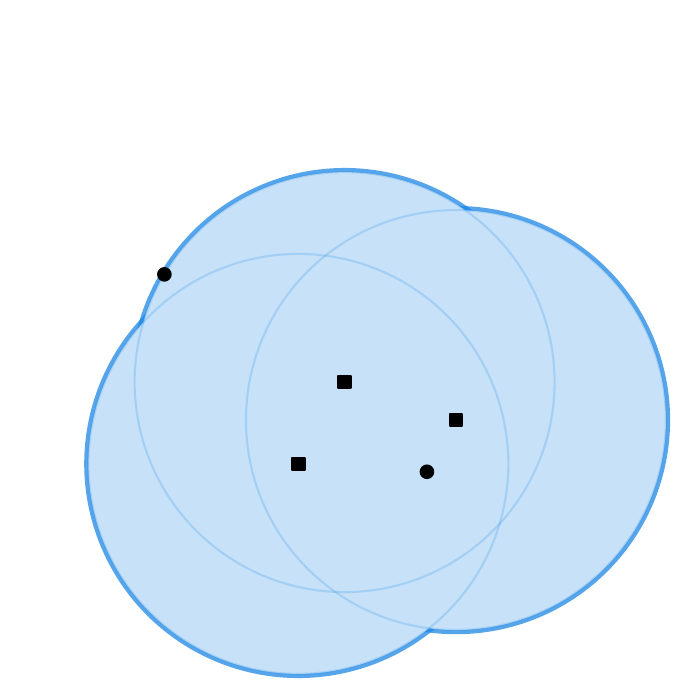}
    \includegraphics[width=0.45\linewidth]{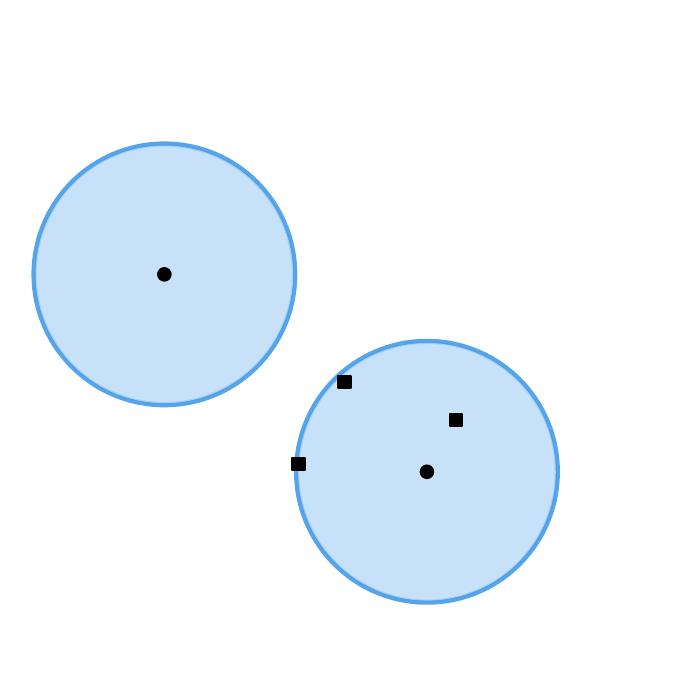}
    \caption{One-sided Hausdorff distances with respect to the Euclidean metric ($d_{Euc}$) between the sets $P$ (black points) and $Q$ (black squares). The left image highlights $d_{Euc}(P,Q)$ and the right image highlights $d_{Euc}(Q,P)$. Clearly, $d_{Euc}(P,Q)>d_{Euc}(Q,P)$.}    
    \label{fig:one-sided_HD}
\end{figure}

The one-sided Hausdorff distance from the set $P$ to $Q$ is the radius of the smallest thickening of $Q$ that contains $P$: 
\begin{align*}
    d(P,Q) = \inf\left\{r\ge0\,:\,P\subseteq\bigcup_{q\in Q}B_d(q;r)\right\},
\end{align*} where $B_d(x;r)$ is the ball of radius $r$ (with respect to the metric $d$) centered at $x$. (We illustrate the two one-sided Hausdorff distances in Figure~\ref{fig:one-sided_HD}.) The Hausdorff distance between $P$ and $Q$ is --- as before --- the maximum of the two radii, $d(P,Q)$ and $d(Q,P)$.
It is a well-known fact that the Hausdorff distance defines a \emph{metric} on the collection of closed, bounded, and nonempty subsets of the metric space $(M,d)$.


\myparagraph{Bregman--Hausdorff divergence.}
To extend the notion of the one-sided Hausdorff distance to the Bregman setting, we use the geometric perspective of thickenings to help us select viable definitions. 

\begin{figure}
    \centering
    \includegraphics[width = 0.45\linewidth]{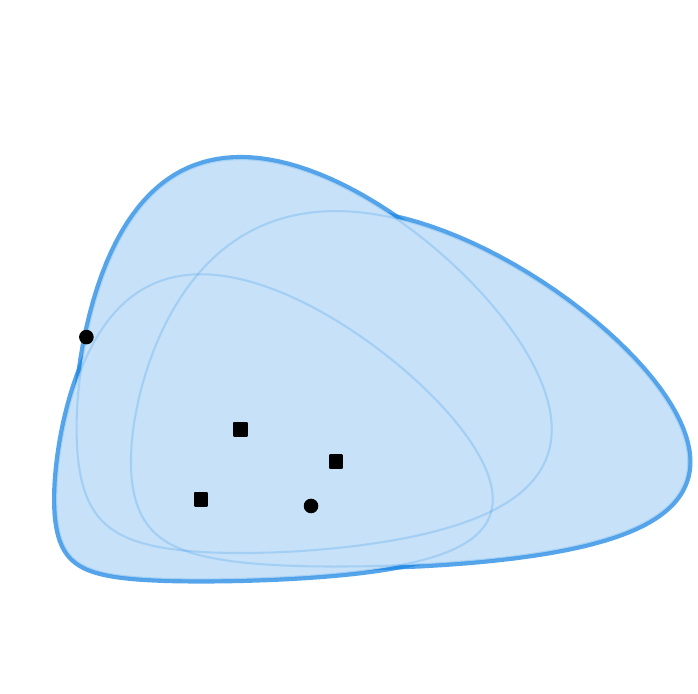}\hfil
    \includegraphics[width = 0.45\linewidth]{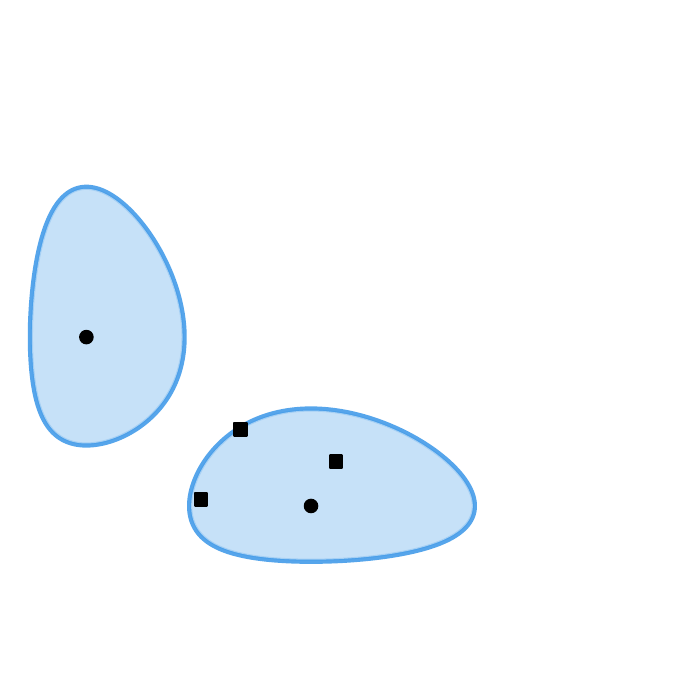}
    \caption{The left portion of the image visualizes the primal Bregman--Hausdorff divergence from $P$ to $Q$, and the right visualizes the Bregman--Hausdorff divergence from $Q$ to $P$. The set $P$ consists of black points, the set $Q$ consists of black squares. Shaded regions are the primal thickening of $Q$ (on the left) and $Q$ (on the right). One sees that the primal thickening of $Q$ has to be of a large radius in order to contain the far left point of $P$. In contrast, the primal thickening of $P$ can contain $Q$ at a smaller radius. Thus, the difference between $H_{KL}(P\|Q)$ and $H_{KL}(Q\|P)$ can be used to see that $Q$ clusters about $P$, but not vice versa.}
    \label{fig:BHExamples}
\end{figure}

Let $F$ be a function of Legendre type, defined on the domain $\Omega$, and let $P$ and $Q$ be two nonempty subsets of $\Omega$. 
The \new{primal} (resp. \new{dual}) \new{thickening} of $Q$ of size $r\ge 0$ is the union of primal (resp. dual) balls of radius $r$, centered at the points in $Q$, with respect to the divergence $D_F$.
We define the \new{primal} (resp. \new{dual}) \new{(one-sided) Bregman--Hausdorff divergence} from $P$ to $Q$, with respect to the divergence $D_F$,  as
\begin{align*}
    H_{D_F}(P\|Q) = \inf\{r\ge0\,:\,P\subseteq \bigcup_{q\in Q}B_{F}(q;r)\},\\
    H'_{D_F}(P\|Q) = \inf\{r\ge0\,:\,P\subseteq \bigcup_{q\in Q}B'_{F}(q;r)\},
\end{align*}
respectively.
See Figure~\ref{fig:BHExamples} for a visualization.

Similarly to the one-sided Hausdorff distances, we have equivalent expressions for both the primal and the dual Bregman--Hausdorff divergence:
\begin{align*}
    H_{D_F}(P\|Q) &= \sup_{p\in P}\inf_{q\in Q}D_F(p\|q),\\
    H'_{D_F}(P\|Q) &= \sup_{p\in P}\inf_{q\in Q} D_F(q\|p).
\end{align*}
These expressions will be useful for computations. It is worth noting that the asymmetry of Bregman divergences allows for more definitions, which however deviate from the natural geometric interpretation of the original Hausdorff distance.

Furthermore, it would be possible to define symmetrized (primal and dual) Bregman-Hausdorff divergence as the maximum of the two variants. However, we refrain from symmetrizing it this way, for the same reason Bregman divergences are typically not symmetrized. Namely, each of the above definitions has a natural interpretation and applications. However, we will introduce a third variant which will be naturally symmetric.

For popular divergences with established names and abbreviations, such as the $KL$ and $IS$ divergences, we shorthand the Bregman--Hausdorff divergences for these divergences to $H_{KL}$ for the $KL$--Hausdorff divergence, and $H_{IS}$ for the $IS$--Hausdorff divergence.


The proposed Bregman--Hausdorff divergences can be used to compare the probabilistic predictions of a machine learning model with a reference model, as we showcase on the example of the $KL$ divergence in the next paragraph. 

\myparagraph{Interpreting the Bregman--Hausdorff divergences with respect to the $KL$ divergence.}
We can  now extend the interpretation of the $KL$ divergence presented in Section~\ref{sec:InfoTheory} to the Bregman--Hausdorff divergences based on this divergence. This new case involves not only pairs of probability vectors --- but pairs of \emph{collections} of such vectors. 

Let $P$ and $Q$ be nonempty collections of probability vectors in $\triangle^{d}$.
If we form a primal $KL$ ball $B_{KL}(q;r)$, with a fixed radius $r\ge 0$, around every point $q\in Q$, then the region covered by these balls will contain all probability vectors that can approximate a vector in $Q$ with an expected loss of at most $r$ bits. 
Now, if the set $P$ is contained in this region, then $r$ is an upper bound on how inefficient the approximation of probabilities in $Q$ is for some vector $p\in P$. Thus, by taking the infimum over all radii such that $P$ is contained within the primal $KL$ balls around $Q$, we can compute how efficient the approximation can be.
The infimum is precisely the primal Bregman-Hausdorff divergence from $P$ to $Q$, $H_{KL}(P\|Q)$; it measures the maximum expected efficiency loss (in bits) if $P$ is used to reasonably approximate $Q$. 
In other words, for any probability vector $p\in P$, there exists a vector $q\in Q$, which $p$ approximates with an expected efficiency loss of at most $H_{KL}(P\|Q)$ bits.

In contrast, the dual Bregman-Hausdorff divergence $H'_{KL}(P\|Q)$ measures the minimum radius for which the dual thickening of $Q$ covers $P$. Each dual ball $B_{KL}'(q;r)$ contains all probability vectors which $q$ approximates with an expected efficiency loss of at most $r$ bits. Thus, when $r=H'_{KL}(P\|Q)$, the union of the dual Bregman balls captures the maximum expected number of bits lost for any probability vector $p\in P$ to be reasonably approximated by a probability vector in $Q$. 

Hence, $H_{KL}(P\|Q)$ and $H_{KL}'(Q\|P)$ measure how $P$ and $Q$ can approximate each other. Specifically, $H_{KL}(P\|Q)$ is the maximum loss of expected bits if any vector $p\in P$ is used to approximate some vector $q\in Q$. On the other hand, for $H'_{KL}(Q\|P)$, every vector $q\in Q$ is approximated by some probability vector in $P$, but not every $q$ will be used as an approximator. For $H_{KL}(P\|Q) = r$, every point in $P$ is contained in some $B_{KL}(q;r)$, so every $p$ is approximating the center of a ball it is contained in. 

We can use the Bregman--Hausdorff divergence in the assessment of performance of machine learning models. Indeed, let $M_1$ and $M_2$ be two different classification models trained using the $KL$ divergence loss (or equivalently the cross entropy loss). We also let $X, Z$ be two datasets, and denote the probabilistic predictions made by the models as $\{M_1(x;\theta_1)\}_{x\in X}$ and $Q=\{M_{2}(z;\theta_2)\}_{z\in Z}$. Then $H_{KL}(P\|Q)$ quantifies a divergence from the set of predictions made by $Q$ towards the set of predictions made by $P$.

We stress that this measurement \emph{does not rely on any explicit pairing} between the outputs of two models (i.e. there is no obvious bijection between $X$ and $Z$). 
Although the data sets $X$ and $Z$ are not explicitly paired, we can still make a reasonable numerical measurement between the two. This is the case when, for example, $M_1=M_2$, and $X$ and $Z$ are training and test data, respectively. In this case, the values $H_{KL}(P\|Q)$  and $H'_{KL}(Q\|P)$ can be used to gauge the generalization power of a model. Importantly, this measurement is consistent with the loss function used to train the model.


\myparagraph{Chernoff--Bregman--Hausdorff distance.} 
We propose one more natural distance measurement. As its name suggests, the Chernoff--Bregman--Hausdorff distance is based on the notion of the Chernoff point. 
As before, we let $F$ be a function of Legendre type, defined on the domain $\Omega$, and let $P$ and $Q$ be two nonempty subsets of $\Omega$. 

\begin{figure}
    \includegraphics[width = .45\linewidth]{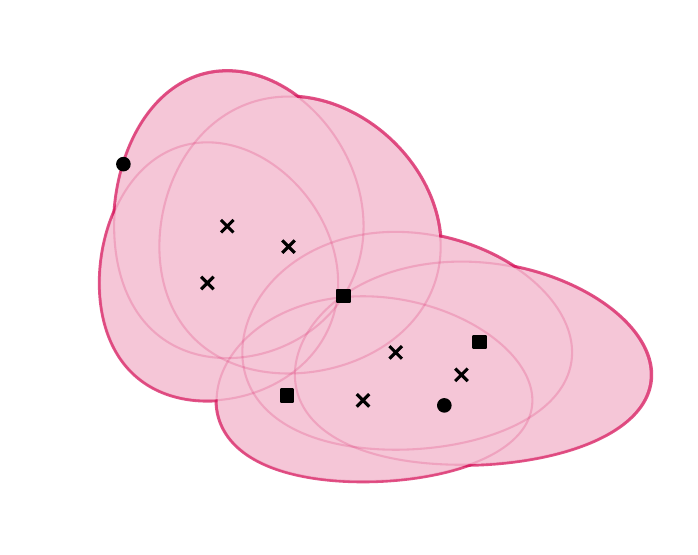}
    \includegraphics[width = .45\linewidth]{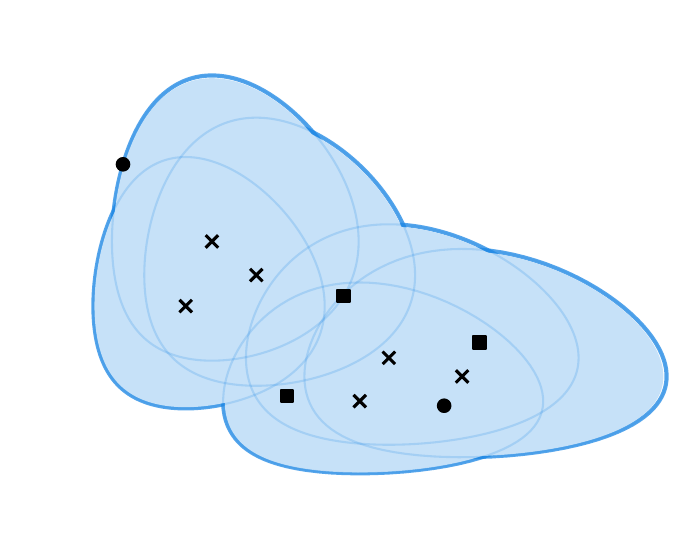}
    \centering
    \caption{On the left, we visualize the primal Chernoff--Bregman--Hausdorff distance between $P$ and $Q$. On the right, we visualize the dual Chernoff--Bregman--Hausdorff distance between $P$ and $Q$. The set $P$ consists of black points, the set $Q$ of black squares. Chernoff points are marked as $\times$. The left is a \textit{dual} thickening of the Chernoff points while the right is a \textit{primal} thickening.}
    \label{fig:CBHExamples}
\end{figure}

For each pair of points $(p,q)\in P\times Q$, write $c_{p,q}$ for the Chernoff point of the set $\{p,q\}$, and write $C = \left\{c_{p,q} \,:\ (p,q)\in P\times Q \right\}$ for the collection of all the Chernoff points. 
Then the primal (resp. dual) Chernoff--Bregman--Hausdorff distance is the smallest size of the (primal resp. dual) thickening of $C$ that contains the union $P\cup Q$.
To be more concrete, we define the \new{primal Chernoff--Bregman--Hausdorff distance} between $P$ and $Q$ as
\begin{align*}
    CH_{D_F}(P,Q) = \inf\{r\ge0\,:\,P\cup Q\subseteq\bigcup_{c\in C}B'_F(c; r)\},
\end{align*}
and the \new{dual Chernoff--Bregman--Hausdorff distance} between $P$ and $Q$ as
\begin{align*}
    CH'_{D_F}(P,Q) = \inf\{r\ge0\,:\,P\cup Q\subseteq\bigcup_{c\in C}B_F(c; r)\}.
\end{align*}
We emphasize that each divergence is named after the type of the Bregman ball that \emph{grows about the pair} $(p,q)\in  P\times Q$, and \emph{not} the balls growing about the Chernoff points.
We visualize the primal and dual Chernoff--Bregman--Hausdorff distances in Figure~\ref{fig:CBHExamples}. Unlike the Bregman--Hausdorff divergences, the Chernoff--Bregman--Hausdorff distance is symmetric. 

If fact, the set $C$ can be viewed as the `average' of $P$ and $Q$ with respect to the chosen divergence at the level of sets. In contrast to symmetrizing the Bregman--Hausdorff divergence by taking the average $(H_{D_{F}}(P\|Q)+H_{D_{F}}(Q\|P))/{2}$, the Chernoff--Bregman--Hausdorff distance avoids mixing directions of divergence computations. In particular, in the context of the $KL$ divergence, the corresponding Chernoff--Bregman--Hausdorff distance inherits the information--theoretical interpretation, as we will see shortly.  In the case of the squared Euclidean distance, $C$ contains the usual arithmetic average $\frac{p+q}{2}$ for each pair $(p,q)\in P\times Q$. 


Again, for the $KL$ and $IS$ divergences, we shorten the notation to $CH_{KL}$ and $CH_{IS}$, respectively.

\myparagraph{Interpreting the Chernoff--Bregman--Hausdorff distance for KL.} As the primal Chernoff--Bregman--Hausdorff distance is defined by taking the infimum of the radius of a dual Bregman ball about the Chernoff points, $CH_{KL}$ gives the least number of expected bits lost using $C$ to approximate both $P$ and $Q$. Similarly, the dual Chernoff--Bregman--Hausdorff distance, where we center primal balls about each Chernoff point, gives the least number of expected bits lost to approximate $C$ by either $P$ or $Q$. 

Returning to machine learning, for models $P=\{M_1(x;\theta)\}_{x\in X}$ and $Q=\{M_2(z;\theta')\}_{z\in Z}$, $C$ can be viewed as the collection of 'average' probability distributions~\cite{ChernoffCompute} for every $(p,q)\in P\times Q$. Therefore, $CH_{KL}(P,Q)$ measures the maximum expected loss of coding efficiency (in bits) when attempting to reasonably approximate both $P$ and $Q$ using $C$. In this sense, $C$ contains `joint approximators' for pairs from $P \times Q$. Similarly, $CH'_{KL}(P, Q)$ measures the maximum expected loss when using $P$ and $Q$ to approximate the set $C$. 
While the Bregman--Hausdorff divergence is applicable if either $P$ or $Q$ is a reference set of vectors, the Chernoff--Bregman--Hausdorff distance is a natural choice when $P$ and $Q$  play the same role.


\section{Algorithms for Bregman--Hausdorff divergences}\label{sec:HausdorffAlgos}
In this section we present first algorithms for the Bregman--Hausdorff divergences. The algorithms rely on the data structure for nearest neighbour search, which we outline next. For the reminder of this section, we let $F$ be a decomposible function of Legendre type, defined on the domain $\Omega$, and let $P$ and $Q$ be two finite, nonempty subsets of $\Omega$. 

\myparagraph{Bregman Kd-trees.} 
We use the Bregman Kd-tree structure and search algorithm mentioned in Section~\ref{sec:BregmanAlgos}. In~\cite{BregmanKdTrees}, Pham and Wagner experimentally show that Kd-trees for Bregman divergences are efficient in a range of practical situations. In particular, they perform better than Cayton's Bregman Ball trees, and alleviate issues with certain other implementations. 

The aforementioned Bregman Kd-tree implementation works for decomposable Bregman divergences~\cite{Decomp_def1, Decomp_def2}, including the squared Euclidean distance, and $KL$ and $IS$ divergences. One additional benefit of this algorithm is its ability to compute the nearest neighbour in either direction.

\myparagraph{Computing the Bregman--Hausdorff divergences.} 
We first provide an algorithm for computing the Bregman--Hausdorff divergences between $P$ and $Q$.

To compute the Bregman--Hausdorff divergence $H_{F}(P\|Q)$ from $P$ to $Q$, we can use a version of the \texttt{Kd-tree} data structure for decomposable Bregman divergences~\cite{BregmanKdTrees}. We first construct a Kd-tree to represent $P$. Then for each $q\in Q$, we search for $\rho^* = \arg\min_{p\in P} D_F(p\|q)$, maintaining value of the largest divergence, $D_F(q\|\rho^*)$.

\begin{algorithm}
\caption{Primal Bregman--Hausdorff divergence algorithm (basic version)} \label{alg:BHKd}
\begin{algorithmic}[1]
\Require Point clouds $P$ and $Q$ of size $n$, $m$; decomposable Bregman divergence $D_F$.
\Ensure Bregman--Hausdorff divergence $H_{D_F}(P\|Q)$
\State $max\_haus \gets 0$
\State $KdTree \gets$ build \texttt{Kd-tree}$(P)$ \label{line:struct}
\For {$q \in Q$}
    \State $nn = KdTree.query(q, D_F)$\Comment{Find $\arg\min_{p\in P}D_F(p\|q)$}\label{line:nnSearch}
    \State $nn\_div = D_F(nn, q)$
    \State $max\_haus = \max(nn\_div, max\_haus)$  
\EndFor
\State \textbf{return} $max\_haus$
\end{algorithmic}
\end{algorithm}
In Algorithm~\ref{alg:BHKd}, we can replace line~\ref{line:struct} and line~\ref{line:nnSearch} with any nearest neighbour structure and search. As the Bregman Kd-tree can be queried for divergences computed in both directions, this algorithm also works for computing the dual Bregman--Hausdorff divergence. For a proximity search algorithm with complexity $O(\mathcal{C}(n, d))$, this algorithm runs in $O(m\cdot \mathcal{C}(n, d))$. 

To approximate the Bregman--Hausdorff divergence, one can use a $(1+\epsilon)$-nearest neighbour algorithm. 

\begin{figure}
   \centering
   \includegraphics[width=0.35\linewidth]{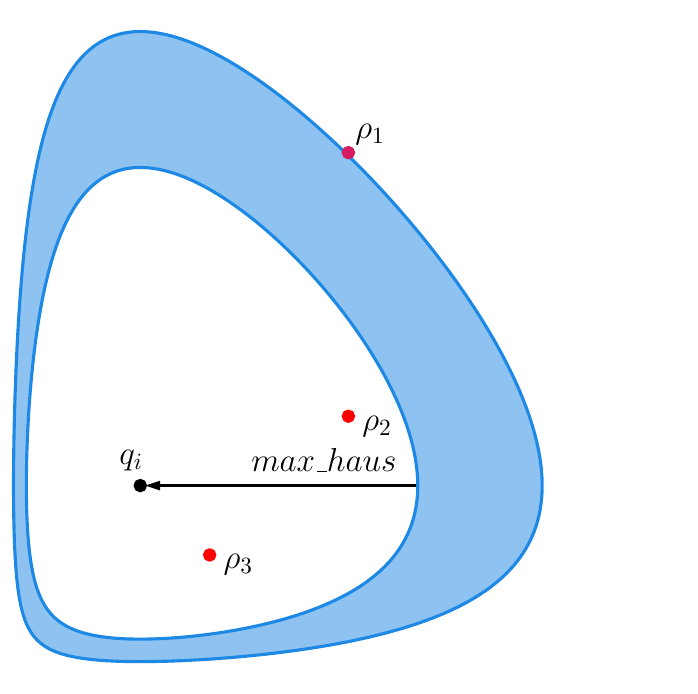}
   \caption{A shrinking shell around a query, $q_i$, whose inner radius is defined by $max\_haus$ and outer radius is defined by $d(\rho_1\|q_i)$. The search will terminate at $\rho_2$, since $D_F(\rho_2\|q) < max\_haus$, instead of returning $\rho_3$ as the true nearest neighbour.}
   \label{fig:BH_ring_alg}
\end{figure}

We can accelerate Algorithm~\ref{alg:BHKd} by adding an early query termination during the Kd-tree search. The the best of our knowledge this technique was introduced to approximate the one-sided Hausdorff distance between 3D meshes~\cite{GutheFastApxHuas}. We adjust the Kd tree query in line~\ref{line:nnSearch} as follows. During each query, candidates for $\arg\min_{p\in P}D_{F}(p\|q_i)$ are found. For any candidate $\rho\in P$, if $D_F(q_i\|\rho) \leq max\_haus$, then we terminate the query. We denote this adjusted search as the $shell\_query$ method on line~\ref{line:shell} in Algorithm~\ref{alg:KdShell}. The $shell$ method which returns the nearest neighbour if the search completes and $Null$ otherwise. An illustration of the above considerations can be seen in Figure~\ref{fig:BH_ring_alg}.
In the worst case scenario, the shell variant will have the same complexity as Algorithm~\ref{alg:BHKd}, but we will see in Section~\ref{sec:Experiments} that reducing the number of complete searches provides significant speed-ups, even in high dimensions.

\begin{algorithm}
\caption{Primal Bregman--Hausdorff divergence shell algorithm (improved)}\label{alg:KdShell}
    \begin{algorithmic}
\Require Point clouds $P$ and $Q$ of size $n$, $m$ respectively. Choice of decomposable Bregman divergence $D_F$.
\Ensure Bregman--Hausdorff divergence $H_{D_F}(P\|Q)$
\State $max\_haus \gets 0$
\State $KdTree \gets$ build \texttt{Kd-tree}$(P)$
\For {$q \in Q$}
    \State $nn = KdTree.shell\_query(q, D_F, max\_haus)$\label{line:shell}
    \If {$nn \neq Null$}
        \State $max\_haus\gets D_F(nn,q)$
    \EndIf
\EndFor
\State \textbf{return} $max\_haus$
    \end{algorithmic}
\end{algorithm}

We will experimentally compare these implementations with a naive algorithm, where the \texttt{Kd-tree} search in line~\ref{line:nnSearch} of Algorithm~\ref{alg:BHKd} is replaced by a linear search.

\myparagraph{Computing the Chernoff--Bregman--Hausdorff distance.}
We also provide an algorithm for the Chernoff--Bregman--Hausdorff distance. To compute the distance, we first determine the set of Chernoff points: $C=\{c_{p,q}\,:\, (p,q)\in P\times Q\}$. 
\begin{algorithm}
\caption{Primal Chernoff--Bregman--Hausdorff divergence algorithm} \label{alg:CHKd}
\begin{algorithmic}
\Require Point clouds $P$ and $Q$ of size $n$, $m$; decomposable Bregman divergence $D_F$.
\Ensure Chernoff--Bregman--Hausdorff divergence $CH_{D_F}(P\|Q)$
\State $max\_haus \gets 0$
\State $C$ =  empty array of size $nm\times \text{dim}(\Omega)$
\State $i = 0$
\For {$q\in Q$}
    \For {$p\in P$}
        \State C[i] = Chernoff$(p,q)$ \Comment{Compute the Chernoff points.}
        \State $i = i+1$
    \EndFor
\EndFor
\State $KdTree \gets$ build \texttt{Kd-tree}$(C)$ \label{line:ChernTree}
\For {$a \in Q\cup P$}
    \State $nn = KdTree.query(a, D_F)$\Comment{Find $\arg\min_{c\in C}D_F(a\|c)$}\label{line:ChernSearch}
    \State $nn\_div = D_F(nn, q)$
    \State $max\_haus = \max(nn\_div, max\_haus)$   
\EndFor
\State \textbf{return} $max\_haus$
\end{algorithmic}
\end{algorithm}

To approximate the Chernoff point $c_{p,q}$ for a pair $(p,q)\in P\times Q$, we perform a bisection search along the line segment connecting $p$ and $q$: $\alpha p + (1-\alpha) q$ with parameter $\alpha \in [0,1] $. This search was proposed by Niesen~\cite{ChernoffCompute}. We assume it runs in time $O(\beta(\epsilon)d)$ to be $\epsilon$ close to the true Chernoff point, with $d$ being the dimension of the ambient space in which the domain $\Omega$ lies.
 Thus this component of the algorithm runs in $O(\beta(\epsilon)dmn)$ time. Letting $O(\mathcal{C}(n+m, d))$ be the complexity of the chosen proximity search algorithm, the search runs in $O(nm\cdot \mathcal{C}(n+m,d))$. This gives us a total running time of $O(mn(\beta(\epsilon)d+\mathcal{C}(n+m, d)))$.

 As with the Bregman--Hausdorff divergence, Kd-trees may be replaced by other exact Bregman nearest neighbour structure and algorithms; an $(1+\epsilon)$ approximation may be used as well.

Computing the dual Chernoff--Bregman--Hausdorff distance requires mapping the input to the Legendre conjugate space. This adds a preprocessing step, but does not affect the complexity of the algorithm. 



\section{Experiments}\label{sec:Experiments}
In this section we compute Bregman--Hausdorff divergences between various data sets. We work both with practical and synthetic data sets, and provide computation times. Generally, we use Algorithms~\ref{alg:BHKd}, \ref{alg:KdShell} with domain $\Omega = \triangle^{d-1}$, where $d$ depends on the chosen data set. We will run experiments using the $KL$ and $IS$ divergences, as well as the squared Euclidean distance. We will focus on Algorithms \ref{alg:BHKd},~\ref{alg:KdShell} because, unlike Algorithm~\ref{alg:CHKd}, they promise to be efficient in practice.

\myparagraph{Compiler and hardware.}
Software was compiled with {Clang 14.0.3}. The experiments were done on a single core of a 3.5 GHz ARM64-based CPU with 4MB L2 cache using 32GB RAM.

\myparagraph{Data sets.} We use predictions from machine learning models and synthetic data sets. We train two neural networks, $M_1$ and $M_2$, on a classification task using CIFAR100, which has 50,000 training images and 10,000 test images. Specifically, we perform transfer learning using EfficientNetB0~\cite{tan2019efficientnet} pretrained on imagenet as a backbone, with ($M_1$) and without ($M_2$) fine-tuning. Both models are trained using the $KL$ divergence as the loss function. The models, $M_1$ and $M_2$, achieve 80.22\%, and 71.74\% test accuracy respectively. From each model, we produce two sets of predictions: $(\text{trn}_{i},\text{tst}_i)$, for $i \in \{1,2\}$. The synthetic data sets are drawn uniformly from the open simplex in dimension 50, 100, and 250. The target dataset, $P$, has 100,000 sample points; the query dataset, $Q$, has 20,000.

\myparagraph{Bregman--Hausdorff computations.} 
One concrete motivation for this measurement was the need to quantify how well one set of predictions approximates another. We set this up as a computation of the Bregman--Hausdorff divergence. We are especially interested in comparing the probabilistic predictions arising from two models of different quality, as well as the training and test data. We were also curious about the difference between the $KL$--Hausdorff divergence and the $IS$--Hausdorff divergence.

In Table~\ref{tab:BregmanHausdorff_divs}, we compare the Bregman--Hausdorff divergence between the four sets of predictions. Divergences change by row, data sets by column, and the units for the $KL$ divergence are in bits.


From the row containing measurements $H_{KL}$ values in Table~\ref{tab:BregmanHausdorff_divs}, the lowest value is $H_{KL}($tst$_1\|$trn$_1)$. This value shows that tst$_1$ predicts trn$_1$ with a maximum expected loss of 1.764 bits. In particular, $H_{KL}(\text{tst}_1\|\text{trn}_1) < H_{KL}(\text{trn}_1\|\text{tst}_1)$ --- the relation we would expect from $M_1$'s predictions on a train and test data sets.
In contrast, $H_{SE}($tst$_1\|$trn$_1) = 0.371$ is the largest value in its row. In particular, $H_{SE}(\text{tst}_1\|\text{trn}_1) > H_{SE}(\text{trn}_1\|\text{tst}_1)$, which is the reverse of how we expect the soft predictions from the training data and test data to behave. On the surface, this would indicate that trn$_1$ is a poor predictor of tst$_1$. However, as the models were trained to minimize the $KL$ divergence, and because of the difference between the Euclidean geometry and the Bregman geometry induced by the $KL$ divergence, values from $H_{SE}$ do not carry the information theoretic interpretation relevant for analysis. This shows the importance of analyzing these models using the proposed $KL$--Hausdorff divergence rather than the standard one-sided Hausdorff distance.

We see a similar effect for the computation of $H_{IS}$ and $H_{IS}'$; the geometry induced by the $IS$ divergence is drastically different from the geometry induced by the $KL$ divergence, and thus the computations lack an obvious interpretation in this context. However, for speech and sound data, where the $IS$ divergence is used for comparisons~\cite{itakura1968analysis} we would expect the ${IS}$--Hausdorff divergence to carry a useful interpretation. We therefore provide the computations to demonstrate that our algorithms can handle various Bregman divergences.
\begin{table}
\centering
    \caption{Bregman--Hausdorff divergences between outputs of two classification models. Values for $H_{KL}$ and $H'_{KL}$ are measured in bits.}
\begin{tabular}{ c r r r r r}\toprule
 & \multicolumn{1}{c}{(tst$_1\|$trn$_1$)}     & \multicolumn{1}{c}{(trn$_1\|$tst$_1$)}
 & \multicolumn{1}{c}{(trn$_1\|$tst$_{2}$)}   &\multicolumn{1}{c}{(trn$_1\|$trn$_2$)}
 & \multicolumn{1}{c}{(trn$_2\|$trn$_1$)}\\
    \cmidrule(lr){2-2}\cmidrule(lr){3-3}\cmidrule(lr){4-4}\cmidrule(lr){5-5}\cmidrule(lr){6-6}
$H_{KL}$  & 1.765b      & 2.215b     & 2.044b & 2.236b & 2.237b \\
$H'_{KL}$ & 3.797b      & 4.541b     & 4.509b & 4.343b & 4.033b \\
$H_{IS}$  & 32,496.887  & 9,822,345.381 & 1,739,646,377.745 & 14,801,113.426 & 584,772.398 \\
$H_{IS}'$ & 3147.685    & 2987.831   & 1998.378 & 2360.0485 & 1309.6230 \\
$H_{SE}$  & 0.371       & 0.296      & 0.243 & 0.234 & 0.271 \\\bottomrule
\end{tabular}
    \label{tab:BregmanHausdorff_divs}
\end{table}
\myparagraph{Timings.}
Finally, we compare the computation speeds of the algorithms using the Bregman Kd-tree (Algorithms~\ref{alg:BHKd},~\ref{alg:KdShell}) and a version that uses a linear search. The results can be seen in Table~\ref{tab:BregmanHausdorff_times}. The left columns are computation times for two sets of predictions. For the Kd-tree algorithm without the shell acceleration, we see significant speed-ups in the left two columns, while the right three columns have a similar time between the two searches. As predictions from classification models tend to cluster near the vertices of the simplex and randomly generated points have more spread across the simplex, we see that the distribution of points heavily influences the speed of computations. Similarly, the run time increases as the dimension increases, which is expected~\cite{kdtree_HiDim, dimcurse}.

In contrast, when we apply early termination via the shell method, we see that we maintain considerable speed-ups even in high dimensions. While the worst case complexity of the shell method is still equivalent to the Kd-tree search, we see up to 1000$\times$ speed-up even in dimension 250.

These speed-ups show that the Bregman--Hausdorff divergence is a practical tool for the comparison of the outputs of machine learning models --- as well as in other situations that require the comparison between two sets of vectors.

\begin{table}
    \caption{Computation times for the $KL$ and $IS$ Hausdorff divergences using the Kd-tree search algorithms and a linear search. Left two columns use data sets from the predictions of $M_1$ and $M_2$ in dimension $100$; the three rightmost columns use synthetic data sets. Speed-up compares the Kd-shell computation times and linear search computation times.}
\begin{tabular}{llrrrrrr}\toprule
                    & & \multicolumn{2}{l}{} & \multicolumn{4}{c}{(Dimension; $P$ size; $Q$ Size)} \\ \cmidrule(lr){5-8}
                        & & \multicolumn{1}{c}{(trn$_{1}\|$tst$_{2})$} & \multicolumn{1}{c}{(trn$_{2}\|$trn$_{1})$} & \multicolumn{1}{c}{\begin{tabular}[c]{@{}c@{}}(10;\\ 100,000;\\ 20,000)\end{tabular}} & \multicolumn{1}{c}{\begin{tabular}[c]{@{}c@{}}(50;\\ 100,000;\\ 20,000)\end{tabular}} & \multicolumn{1}{c}{\begin{tabular}[c]{@{}c@{}}(100;\\ 100,000;\\ 20,000)\end{tabular}} & \multicolumn{1}{c}{\begin{tabular}[c]{@{}c@{}}(250;\\ 100,000;\\ 20,000)\end{tabular}} \\ \cmidrule(lr){3-3}\cmidrule(lr){4-4}\cmidrule(lr){5-5}\cmidrule(lr){6-6}\cmidrule(lr){7-7}\cmidrule(lr){8-8}
\multirow{3}{*}{\rotatebox{90}{K\,L}} 
                    & Kd-shell  & 0.475s    & 1.635s   & 0.023s    & 0.782s    & 1.361s    & 2.195s   \\
                    & Kd-tree   & 1.741s    & 6.796s    & 2.431s   & 481.151s  & 999.262s  & 2413.94s \\
                    & Linear    & 284.307s  & 1417.320s & 112.844s & 563.170s  & 1113.630s & 2753.32s \\
                    \cmidrule(lr){3-3}\cmidrule(lr){4-4}\cmidrule(lr){5-5}\cmidrule(lr){6-6}\cmidrule(lr){7-7} \cmidrule(lr){8-8}
                & Speed-up  &  598.541$\times$ & 866.862$\times$  &  4906.261$\times$&  720.166$\times$ & 818.244$\times$ & 1254.360$\times$\\\midrule\midrule
\multirow{3}{*}{\rotatebox{90}{I\,S}} 
                    & Kd-shell  & 0.209s    & 1.012s    & 0.018s& 0.666s    & 1.487s    & 3.921s \\
                    & Kd-tree   & 7.091s    & 23.921s   & 3.850s   & 412.387s  & 738.372s  & 1645.400s \\
                    & Linear    & 279.703s  & 1391.390s & 115.124s & 574.264s  & 1129.01s  & 2803.430s \\
                    \cmidrule(lr){3-3}\cmidrule(lr){4-4}\cmidrule(lr){5-5}\cmidrule(lr){6-6}\cmidrule(lr){7-7}\cmidrule(lr){8-8}
                & Speed-up  & 1338.292$\times$ &  1374.891$\times$ & 6395.778$\times$ & 862.258$\times$ & 759.254$\times$ & 714.978$\times$ \\\bottomrule
\end{tabular}
    \label{tab:BregmanHausdorff_times}
\end{table}

\section{Conclusion}\label{sec:Ending}
In this paper, we outlined the field of Bregman geometry and extended a common geometric tool to this new setup. Specifically, we 
defined several variants of Bregman--Hausdorff divergences and highlighted situations in which they can be meaningfully used. In particular, when the underlying pairwise divergence is the Kullback--Leibler divergence, its Hausdorff counterpart has a clear information-theoretical interpretation.

We also outlined novel algorithms for computing these divergences. Our experiments show that these algorithms are efficient in practice. In particular, a careful application of the recently developed Bregman Kd-trees results in several orders of magnitude speed-up compared to a basic algorithm. This is surprising, given that the Kd-trees are known to perform poorly in such high-dimensional situations --- however the special structure of the Bregman--Hausdorff computations allows for more efficient computations. Understanding this aspect of our work is an interesting future direction.

Overall, we hope that the Bregman--Hausdorff divergence will be a valuable tool, and in particular integrate into machine learning pipelines which frequently rely on Bregman divergences.


\bibliographystyle{plainurl}
{
\bibliography{main}
}
\end{document}